\documentclass[twoside]{article}

\usepackage[preprint]{aistats2026}
\usepackage[round]{natbib}

\usepackage[utf8]{inputenc} 
\usepackage[T1]{fontenc}    
\usepackage{hyperref}       
\usepackage{url}            
\usepackage{booktabs}       
\usepackage{amsfonts}       
\usepackage{nicefrac}       
\usepackage{microtype}      
\usepackage[table]{xcolor}         
\usepackage{multirow}
\usepackage{makecell}
\usepackage{amsmath,amsthm}
\usepackage{amssymb}
\usepackage{graphicx}
\usepackage{subcaption}
\usepackage{enumitem}
\usepackage{changepage}
\usepackage{cleveref}
\usepackage[normalem]{ulem}
\usepackage{algorithm} 
\usepackage{algpseudocode}
\algrenewcommand{\algorithmiccomment}[1]{\hfill\textnormal{\tiny(#1)}}

\DeclareMathOperator*{\argmax}{arg\,max}

\newtheorem{assumption}{Assumption}

\usepackage[colorinlistoftodos,prependcaption]{todonotes}
\newcommand{\bp}{\mathbf{p}}
\newcommand{\br}{\mathbf{r}}
\newcommand{\bx}{\mathbf{x}}
\newcommand{\bw}{\mathbf{w}}

\usepackage[dvipsnames]{xcolor}
\usepackage[most]{tcolorbox} 
\tcbset{enhanced,breakable,boxsep=1mm,left=1.5mm,right=1.5mm,top=1mm,bottom=1mm}

\newtcolorbox{promptbox}[1][]{%
  colback=RoyalBlue!3,          
  colframe=RoyalBlue!60,        
  colbacktitle=black!8,     
  coltitle=black,           
  fonttitle=\bfseries,
  arc=2mm,                  
  attach boxed title to top left={xshift=2mm,yshift*=-2mm},
  boxed title style={colframe=black!30},
  #1                         
}

\begin{document}

%
\runningtitle{ProxRouter: Proximity-Weighted LLM Query Routing for Improved Robustness to Outliers}

%
\runningauthor{Shivam Patel, Neharika Jali, Ankur Mallick, Gauri Joshi}

\twocolumn[
 
\aistatstitle{ProxRouter: Proximity-Weighted LLM Query Routing \\for Improved Robustness to Outliers}

\aistatsauthor{ Shivam Patel\textsuperscript{1}\textsuperscript{\textdagger} \And Neharika Jali\textsuperscript{1} \And Ankur Mallick\textsuperscript{2} \And Gauri Joshi\textsuperscript{1} }

\aistatsaddress{\textsuperscript{1}Carnegie Mellon University,  \textsuperscript{2}Microsoft} ]

\begingroup
\renewcommand\thefootnote{\textdagger}        
\footnotetext{Corresponding author}
\endgroup

\begin{abstract}

Large language model (LLM) query routers are critical to modern AI platforms as they seek to improve efficiency by assigning inference queries to accurate, yet low-cost models. Parametric routers typically use trained neural networks for LLM selection but suffer from retraining and maintenance overheads. Nonparametric routers are training-free, instead estimating LLM accuracy and cost via similarity between encodings of the input query and training set queries. However, like their parametric counterparts, nonparametric routers struggle to generalize to outlier queries, an issue exacerbated by limited diversity in training sets which are costly to expand and difficult to keep current with ever-evolving use cases. We propose ProxRouter, which applies an exponentially tilted aggregation mechanism to balance bias and variance in nonparametric routers, improving their robustness to outliers. Experiments show ProxRouter enhances outlier routing while preserving inlier performance with minimal overhead.
\end{abstract}

\section{Introduction}\label{sec:introduction}

Generative AI, particularly large language models (LLMs) \cite{NEURIPS2020_1457c0d6_lm_are_few_shot_learners,NIPS2017_3f5ee243_attention_isallyouneed}, has achieved remarkable breakthroughs, surpassing human-level accuracy on a wide range of tasks \cite{openai2024gpt4technicalreport}. Yet, state-of-the-art models contain billions or even trillions of parameters \cite{kaplan2020scalinglawsneurallanguage}, making inference excessively costly. As LLMs are integrated into an expanding set of applications, the landscape has rapidly diversified, with models of varying sizes, costs, and levels of specialization. The open-source ecosystem alone hosts hundreds of thousands of pretrained and fine-tuned models\footnote{As of this writing, over $282{,}000$ models are available on \href{https://huggingface.co/models}{huggingface.co/models}.}, spanning a broad spectrum of capabilities \cite{hu2022lora}. Using a frontier-scale model for every query is often unnecessary, particularly when smaller, cheaper models can deliver equally accurate responses to simpler inputs. In fact, domain-specific lightweight models frequently outperform general-purpose LLMs on specialized downstream tasks while operating at a fraction of the cost \cite{touvron2023llamaopenefficientfoundation}. This combination of model abundance and the high expense of large-scale inference has driven growing interest in \emph{efficient query routing}, which seeks to dynamically select the most appropriate model to deliver accurate responses at minimal cost.

\begin{figure}
    \centering
    \includegraphics[width=0.7\linewidth]{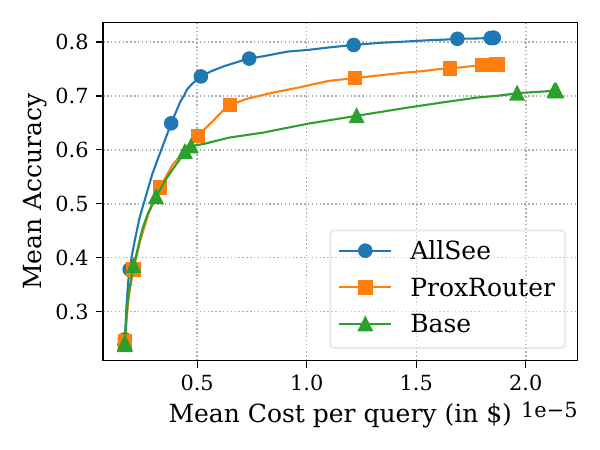}
    \caption{\small The \emph{Base} router (a nearest neighbors router for this plot), which is only trained on inlier queries, fails to generalize to outlier tasks at test time, showing upto $15\%$ average accuracy drop for a given cost, relative to the \emph{AllSee} router trained on both inliers and outliers. Our proposed \emph{ProxRouter}, although trained only on inlier queries, improves robustness to outliers and achieves a better accuracy-cost trade-off (experimental details in \Cref{sec:experiments}).}
    \label{fig:ood_problem_firstpage}
\end{figure}

\begin{figure}[t]
  \centering
  \begin{minipage}[t]{0.55\linewidth}
    \centering
    \includegraphics[width=\linewidth]{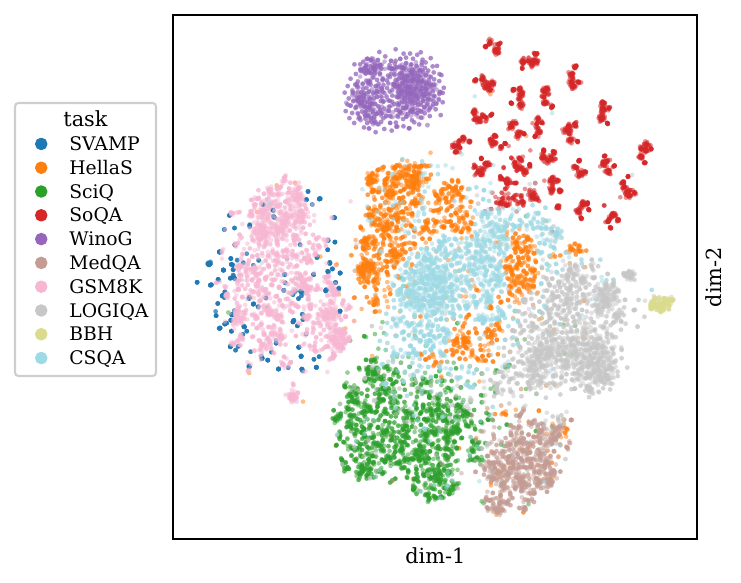}
  \end{minipage}
  \hfill
  \begin{minipage}[t]{0.43\linewidth}
    \centering
    \includegraphics[width=\linewidth]{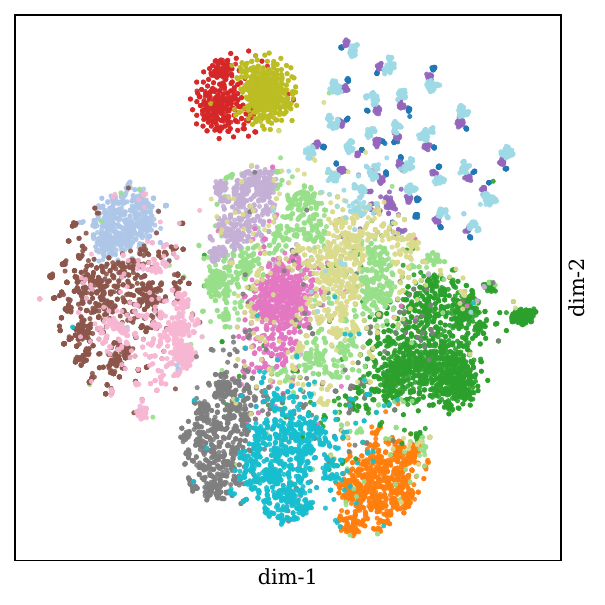}
  \end{minipage}
  \caption{\small High-dimensional query encodings downprojected using t-SNE \cite{tsne}. \textit{Left}: colored by task. \textit{Right}: colored by cluster assignment through $K$Means clustering ($K=16$). Queries from same task occupy compact, localized neighborhoods in encoding space, allowing clustering to recover semantically coherent regions aligned with query types.}
  \label{fig:encoding_queries}
\end{figure}

Query routing seeks to identify the most suitable model from a large pool of LLMs, minimizing inference costs while maintaining response quality. A central challenge is estimating the accuracy and cost of responses to unseen queries. Most methods begin with fixed-dimensional query encodings generated by deterministic sentence encoder models \cite{cer2018universalsentenceencoder, paraphrase_albert_reimers-2019-sentence-bert}, which form the backbone of both parametric and nonparametric routers. Parametric routers train small neural networks (e.g., multi-layer perceptrons) to predict correctness or cost \cite{hu2024routerbench,ding2024hybridllmcostefficientqualityaware, ding2025bestrouteadaptivellmrouting, zhuang2024embedllmlearningcompactrepresentations, sakota_2024_flyswat,huang2025routerevalcomprehensivebenchmarkrouting}, or leverage ranking and reward models \cite{ong2025routellmlearningroutellms, treacle_thrifty_reasoning, chen2025tagrouterlearningroutellms}. In contrast, nonparametric routers are training-free and operate directly in the query encoding space. Two canonical forms dominate: clustering-based approaches (e.g., $K$Means) \cite{zhang2025avengerssimplerecipeuniting, avengers_pro_zhang2025gpt5makingllmscheaper, jitkrittum2025universal, hu2024routerbench} and nearest-neighbor methods (e.g., $k$NN) \cite{li2025rethinkingpredictivemodelingllm, zhuang2024embedllmlearningcompactrepresentations, jitkrittum2025universalmodelroutingefficient, stripelis2024tensoroperaroutermultimodelrouter}. Despite their simplicity, nonparametric routers often rival or surpass parametric ones, likely due to the expressive power of encoder-based query embeddings (\Cref{fig:encoding_queries}), which make the mapping from queries to model abilities straightforward. Moreover, clustering and nearest-neighbor methods naturally accommodate new queries and models without retraining, a key advantage over parametric designs. \emph{For these reasons, in this work, we focus on the nonparametric routing paradigm.}

The success of query routers depends critically on the quality of their accuracy and cost estimates for incoming queries. High-quality estimates require large and diverse router training datasets in which queries are evaluated on all, or a substantial subset of models in the pool. Running inference on all models for each query is expensive, and thus, routers are typically trained only for a limited set of downstream tasks. As new applications emerge, entirely novel query types (e.g., new programming languages) can lead to poor routing performance (see \Cref{fig:ood_problem_firstpage}). Even minor changes in phrasing, for instance, appending Chain-of-Thought instructions \cite{wei2023chainofthoughtpromptingelicitsreasoning}, can shift both accuracy and cost, further complicating generalization. Retraining the router on these outlier queries will require evaluating those queries on all models in the pool, which can be prohibitively expensive. While some approaches replace encoder-based routers with LLMs (e.g., GPT-4o-mini) to make routing decisions directly, the resulting increase in routing cost is impractical at scale. Thus, although accurate predictions of model correctness across all queries remain the objective, the central challenge lies in achieving robust generalization to outlier and unseen queries. \Cref{fig:ood_problem_firstpage} illustrates how poor generalization to a diverse set of queries leads to performance drops, underscoring the need for better robustness.

\begin{figure}[t]
  \centering
    \includegraphics[width=\linewidth]{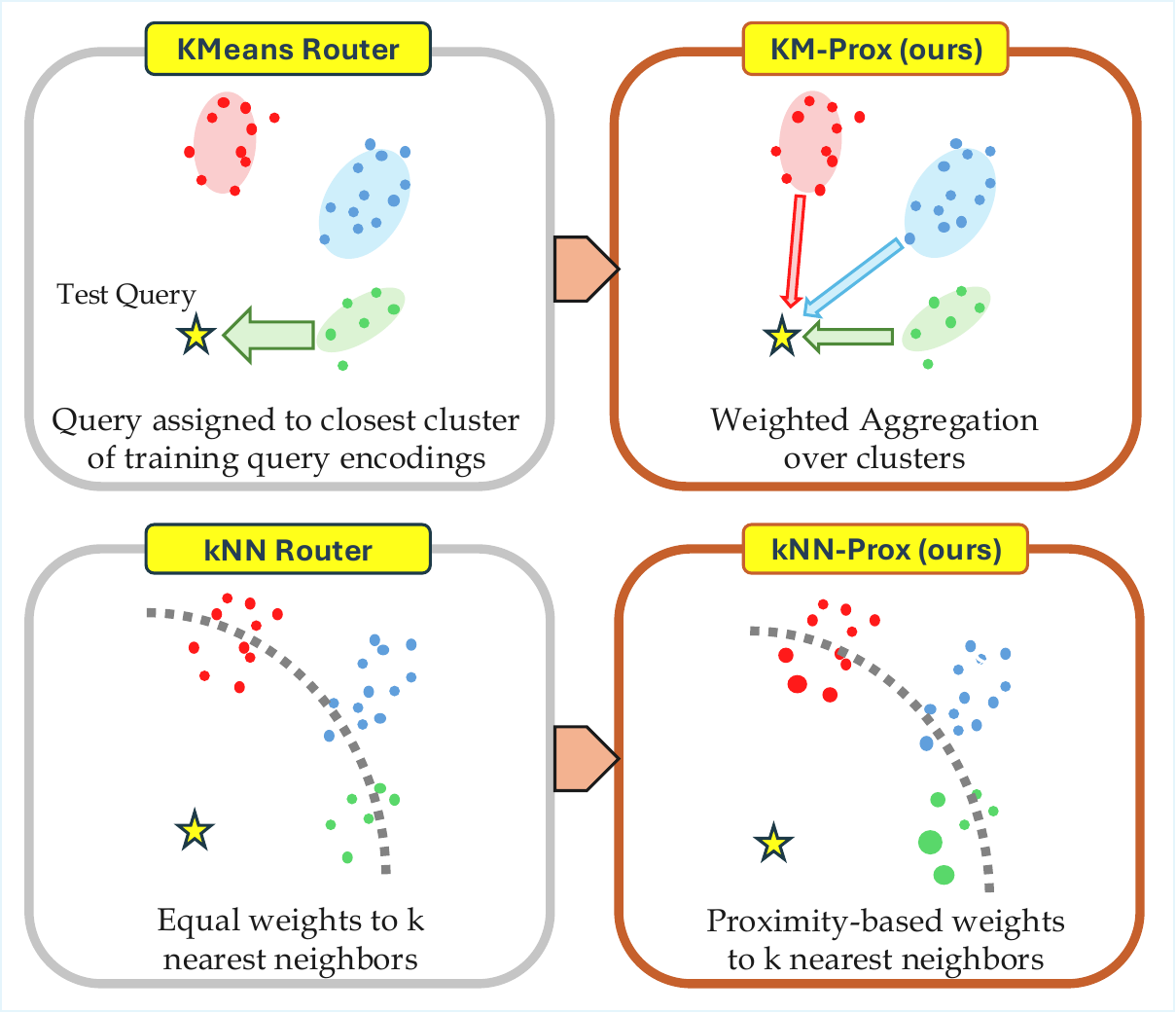}
  \caption{\small Comparison of our proximity-weighted $K$Means clustering and $k$NN routers. While $K$Means assigns a test query encoding to the closest cluster of training query encodings to obtain the accuracy and cost estimates, $K$M-Prox takes a proximity-weighted combination of the $K$ clusters' for each test query. Similarly, $k$NN-Prox takes a proximity-weighted combination of the estimates of the $k$ nearest neighbors of the test query.}
  \label{fig:high_level_illustration}
\end{figure}

In this work, we propose ProxRouter, a method that enhances the robustness of clustering- and nearest-neighbor–based nonparametric routers to outlier queries by improving the model accuracy and cost estimates for all queries. The main contributions of our paper are summarized below:
\vspace*{-8pt}
\begin{itemize}\setlength{\itemsep}{-1pt}
\item In \Cref{sec:problem_formulation}, we develop a unified framework for nonparametric routers, of which clustering and nearest-neighbor routers are special cases. The framework formulates accuracy and cost estimates as a weighted average of estimates provided by reference clusters or neighbors. This enables standardized design and analysis of more general proximity-based routers.
\item In \Cref{sec:proxrouter}, we introduce ProxRouter, which generalizes clustering and nearest-neighbor routers into proximity-weighted versions, where clusters or neighbors closer to a test query receive higher weights (as illustrated in \Cref{fig:high_level_illustration}). Aggregation weights are set according to a minimum-variance prior and exponentially tilted to reduce bias. By using soft aggregation over training data, ProxRouter yields more accurate model performance and cost estimates, particularly for outlier queries. ProxRouter does not require any outlier detection and its associated overhead, since it applies the soft aggregation to all queries, both inliers and outliers.
\item In \Cref{sec:experiments}, we extensively evaluate ProxRouter on a diverse set of $14$ LLMs and queries from $10$ publicly available datasets with numerous outlier-task variations. Results demonstrate that ProxRouter significantly improves robustness to outlier queries while preserving inlier performance. It increases the area under the accuracy–cost curve (AUC), bringing it closer to that of an AllSee router trained on both inlier and outlier queries.
\end{itemize}
Finally, in \Cref{sec:conclusions}, we summarize our findings and discuss extensions of the framework to other nonparametric routers. Details of the bias–variance decomposition, more detailed related work, and additional experimental results are deferred to the Appendix. 

\section{Problem Formulation} \label{sec:problem_formulation}

\subsection{Objective of the Query Router}
\label{subsec:routing_objective}
The typical query router's objective is to maximize the accuracy of responses to inference queries subject to cost constraints, or to minimize cost subject to accuracy constraints.  We consider a Lagrangian relaxation (\cite{jitkrittum2025universalmodelroutingefficient, hu2024routerbench, avengers_pro_zhang2025gpt5makingllmscheaper, li2025rethinkingpredictivemodelingllm, li2025llmbanditcostefficientllm}) of the constrained formulation, which results in an unconstrained objective:
\begin{equation}\label{eq:routing_objective}
    m^*(\bx) \leftarrow \argmax_{m \in \mathcal{M}}[\;\text{acc}^{(m)}(\bx) - \lambda\cdot\text{cost}^{(m)}(\bx)\;]
\end{equation}
where $\{\text{acc}^{(m)}(\bx), \text{cost}^{(m)}(\bx)\}$ denote response accuracy and cost incurred by models $m\in \mathcal{M}$ for the given test query denoted by its sentence encoding $\bx\in\mathbb{R}^{d_\text{enc}}$, which is generated by an encoder language model \footnote{We refer to language queries through their sentence encoding \cite{cer2018universalsentenceencoder} representations $\bx\in\mathbb{R}^{d_\text{enc}}$ for our analysis, as it provides a fixed-dimensional representation of the abstract space of language queries.} \cite{cer2018universalsentenceencoder}. The Lagrangian parameter $\lambda$ controls the relative importance the router gives to accuracy versus the cost of the generated responses. 

We denote the scalar objective value $ U^{(m)}(\cdot) :\mathbb{R}^{d_\text{enc}}\rightarrow\mathbb{R}$ corresponding to any model $m$ given query $\bx$ as 
\begin{align}
   U^{(m)}(\bx) &= \text{acc}^{(m)}(\bx) - \lambda\cdot\text{cost}^{(m)}(\bx) \\ 
   &= \overline{U}^{(m)}(\bx) + \epsilon^{(m)}(\bx)
\end{align}
where we decompose the objective into its expected value $\overline{U}^{(m)}(\bx) = \mathbb{E}[U^{(m)}(\bx)]$ for given query $\bx$, and $\epsilon^{(m)}(\bx)$ a zero mean noise that captures the effect of generation randomness due to the stochastic nature of language models leading to different response correctness and response lengths. Since the stochastic noise is difficult to control or estimate, we reframe the query routing problem as follows. 
\begin{align}
m^*(\bx) &\leftarrow \argmax_{m\in\mathcal{M}} \overline{U}^{(m)}(\bx) \label{eq:routing_obj_mean}   \\ 
&\approx \argmax_{m\in\mathcal{M}} \widehat{U}^{(m)}(\bx)  \label{eq:routing_obj_hat}
\end{align}

Thus, the main task of the router is to produce estimates $\widehat{U}^{(m)}(\bx)$ of $\overline{U}^{(m)}(\bx)$ for each incoming query $\bx$. A mismatch between training queries and test queries leads to poor estimation of $\widehat{U}^{(m)}(\bx)$ and eventually misrouting through \cref{eq:routing_obj_hat}.

Existing nonparametric routers such as $K$Means and $k$NN produce these estimates using the nearest cluster of training queries or the $k$ nearest neighboring training queries. For outlier queries that significantly differ from training queries, $K$Means and $k$NN routers also produce poor estimates $\widehat{U}^{(m)}(\bx)$, resulting in suboptimal routing decisions. Below, we present a general framework that subsumes these existing routers, and lays the foundation for our proposed ProxRouter, which is more robust to outliers.

\subsection{A Unified Representation of Clustering and Nearest Neighbors Routers}\label{subsec:general_form}

For a training set of queries $\mathcal{X}_\text{tr} \subseteq \mathcal{X}$ in the space of all possible queries $\mathcal{X}$, we have model accuracy and cost observations from previously generated responses. The encoding space projections $\{\bx_\text{tr}: \bx_\text{tr}\in\mathcal{X}_\text{tr}\}$ of the training queries (denoted by $\bx_\text{tr}$) forms the basis of all nonparametric routers. We introduce the notion of a reference set $\mathcal{I}$ which represents the sources of information used by the estimator $\widehat{U}^{(m)}(\bx)$ for any test query $\bx$. Each element $\{i:(\br_i, V^{(m)}_i)\}$  in the reference set is indexed by $i\in \{1,\cdots,|\mathcal{I}|\}$, and has two associated attributes: the reference vector $\br_i\in\mathbb{R}^{d_\text{enc}}$ in the encoding space, and $V_i^{(m)}\in\mathbb{R}$ denoting the objective value. 

For clustering-based routers (eg. $K$Means router), the reference set $\mathcal{I}$ corresponds to the set of clusters ($|\mathcal{I}| = K$, number of clusters). The number of queries in each cluster $c_i$ for $i\in \{1,\cdots,|\mathcal{I}|\}$ is denoted by $n_i$. The reference vector $\br_i = \sum_{\bx_\text{tr}\in c_i}{\bx_\text{tr}}/n_i$ is the geometric centroid of training query encodings in the cluster $c_i$. The model objective value $V_i^{(m)} = \sum_{\bx_\text{tr}\in c_i} U^{(m)}(\bx_\text{tr})/n_i$ is the average of the observed model objective values for training queries in the cluster $c_i$, where $U^{(m)}(\bx_\text{tr}) = \text{acc}^{(m)}(\bx_\text{tr}) - \lambda\cdot\text{cost}^{(m)}(\bx_\text{tr})$ for a fixed $\lambda$. 

For nearest neighbor based routers (eg. $k$NN router), each training query $\bx_\text{tr}\in\mathcal{X}_\text{tr}$ corresponds to unique element in reference set $\mathcal{I}$ ($|\mathcal{I}|=|\mathcal{X}_\text{tr}|$, the number of queries in training set). The representative vector $\mathbf{r}_i = \bx_\text{tr}$ is the encoding vector for training query itself, and the model objective value $V^{(m)}_i = U^{(m)}(\bx_\text{tr}) = \text{acc}^{(m)}(\bx_\text{tr}) - \lambda\cdot\text{cost}^{(m)}(\bx_\text{tr})$ for query $\bx_{\text{tr}}$. 

Using the reference set notation above, we denote the family of estimators of $\overline{U}^{(m)}(\bx)$ for any test query $\bx$ as
\begin{equation}\label{eq:general_router_form}
    \widehat{U}^{(m)}(\bx) = \sum_{i \in [|\mathcal{I}|]} w_i(\bx) V^{(m)}_i 
\end{equation}
where $w_i(\bx)\geq 0, \;\;\sum_{i \in [|\mathcal{I}|]} w_i(\bx) = 1$. 

\paragraph{$K$Means and $k$NN routers as Special Cases} The baseline $K$Means router allocates weight $w_i(\bx) = 1$ for the index $i$ corresponding to the closest cluster $c^*$ to test query $\bx$ and $w_i(\bx) = 0$ to all other clusters. The $k$NN router allocates $w_i(\bx)=1/k$ for the $k$ nearest neighbors of query $\bx$ and $w_i(\bx)=0$ to all other reference points. The distance $d(\mathbf{x},\mathbf{r}_i)$ between the test query encoding $\bx$ and reference vectors $\mathbf{r}_i$ is calculated by an appropriate measure $d(\cdot,\cdot):\mathbb{R}^{d_\text{enc}}\times\mathbb{R}^{d_\text{enc}} \rightarrow \mathbb{R}$ such as cosine distance, Euclidean distance etc. whose choice is guided by the design of the deterministic encoder model. We utilize cosine distance in our experimental analysis, as it is suitable for a variety of sentence encoders.

\paragraph{Drawbacks of Existing Techniques for Oulier Queries} For $K$Means router, closest cluster assignment presents a hard decision boundary, and thus, slight changes in the test query $\bx$ lead to abrupt changes in the predicted objective values in \cref{eq:general_router_form}, especially for outlier queries that are well outside the clusters. Moreover, $K$Means clustering is agnostic to intracluster spread, which is indicative of the variance $\text{Var}[V^{(m)}_i]$ of model objectives for individual reference set elements indexed by $i\in \{1,\cdots, |\mathcal{I}| \}$. In $k$NN routers, outlier queries whose $k$ nearest neighbors lie far away suffer under uniform weighting as distant neighbors receive the same weight as closer ones, degrading the estimate. These issues adversely impact routing performance on outlier queries, which we seek to address via our proposed proximity-weighted router described below.

\section{ProxRouter: Proximity Weighted Aggregation Router} \label{sec:proxrouter}

To improve the model accuracy and cost estimates produced by the query router for outlier queries while preserving inlier performance, our proposed method ProxRouter (\Cref{alg:proxrouter}) defines new aggregation weights in \cref{eq:general_router_form}, and applies them to all test queries (both inliers and outliers) alike, thereby eliminating any need for inlier-vs-outlier classification for a test query.  It initializes the aggregation weights in \cref{eq:general_router_form} as minimum variance priors for test query $\bx$ (which we refer to as $\bp (\bx)=[p_1(\bx), \cdots, p_{|\mathcal{I}|}(\bx)]$) over the reference set, and then exponentially tilts them for each reference element based on proximity to the query $\bx$ to obtain bias-controlled weights $\bw (\bx)=[w_1(\bx), \cdots, w_{|\mathcal{I}|}(\bx)]$. This two-stage design approach, explained in detail below, results in better estimates of the objective values $\widehat{U}^{(m)}(\bx)$ and improved routing decisions. While we focus our attention on improving the robustness of the standard $K$Means and $k$NN-based query routers due to their widespread adoption and interpretability, more advanced techniques such as fuzzy, spectral, and kernel methods have been studied in statistics literature \cite{FERRARO2024110_fuzzy_kmeans, NIPS2001_801272ee_spectral_clustering, WandJones1995_kernel_smoothing,Berger1985_bayesian}. Future work may consider extending this approach to more advanced statistical methods.

\begin{algorithm}[htb]
  \caption{ProxRouter} 
  \label{alg:proxrouter}
  \centering
  \begin{minipage}{0.99\linewidth}

\begin{algorithmic}[1]
\State \textbf{Input:} Test query $\bx$
\State \textbf{Given:} Training queries $\mathcal{X}_\text{tr}$, LLM pool $\mathcal{M}$, and Reference set elements $\{i: (\br_i, V^{(m)}_i)\} \;\forall\; i \in [|\mathcal{I}|]$. 

\State \textbf{Objective:} Find $\argmax_{m\in\mathcal{M}} U^{(m)}(\bx)$. 
\State \textbf{Procedure:}
\State \hspace{1em} $p_i(\bx) \propto [\text{Var}(V^{(m)}_i)]^{-1} \Comment{least variance priors}$
\State \hspace{1em} $w_i(\bx) \propto p_i(\bx) \exp(-\phi_i(\bx)/\tau) \Comment{proximity tilting}$
\State \hspace{1em} $\widehat{U}^{(m)}(\bx) \leftarrow \sum_{i} w_i(\bx)\,V^{(m)}_i \Comment{estimated objective}$
\State \hspace{1em} $m^\star(\bx) \leftarrow \arg\max_{m\in\mathcal{M}} \widehat{U}^{(m)}(\bx) \Comment{chosen best model}$
\State \textbf{Return:} $m^\star(\bx)$.
\end{algorithmic}

  \end{minipage}
\end{algorithm}

\subsection{Least Variance Priors} 
\label{subsec:min_var_prior}
When choosing the aggregation weights $w_i(\bx)$ in the estimate $\widehat{U}^{(m)}(\bx)$ as per \cref{eq:general_router_form}, one must consider the heterogeneous variance levels $\text{Var}[V^{(m)}_i]$ for elements in the reference set. The optimal least variance estimator weights (priors) $\bp(\bx)$ are of the form of $p_i(\bx)\propto (\text{Var}[V^{(m)}_i])^{-1}$. The proof involves minimizing the variance of RHS in \cref{eq:general_router_form} subject to the weights summing to unity, and it is detailed in \Cref{app:min_var_prior}. 

For the $K$Means clustering router, the variance of each cluster summary $\text{Var}[V_i^{(m)}]$ can be expressed as the sum of variances of the $n_i$ training query points ($\bx_\text{tr} \in c_i$) belonging to cluster $c_i$, and under the independence assumption (see \Cref{app:bias_variance_tradeoff}):
\begin{align}
\text{Var}[V_i^{(m)}] &= \mathbb{E}\Bigg[ \frac{1}{n_i^2}\sum_{\bx_\text{tr}\in c_i} \big(\epsilon^{(m)}(\bx_\text{tr})\big)^2 \Bigg] \nonumber \\
&= \frac{1}{n_i^2}\sum_{\bx_\text{tr}\in c_i} \text{Var}\left[\epsilon^{(m)}(\bx_\text{tr})\right]. \label{eq:cluster_variance}
\end{align}

Since the exact per-query variance $\text{Var}[\epsilon^{(m)}(\bx_\text{tr})]$ is unknown in practice, we estimate $\text{Var}[V_i^{(m)}]$ based on heuristics derived from query similarity in the encoding space. Intuitively, clusters that are more geometrically dispersed contain semantically diverse queries, implying that $\text{Var}[V_i^{(m)}]$ within such clusters is larger than in compact clusters. Formally, we define the intra-cluster “spread” as $s_i = \sum_{\bx_\text{tr} \in c_i} d(\bx_\text{tr},\br_i)/n_i$, the mean distance of queries in $c_i$ from their centroid $\br_i$. Consequently, $\text{Var}[V^{(m)}_i]$ grows with $s_i$ but decreases with $n_i$ (as suggested by \cref{eq:cluster_variance}, where variance accumulates linearly in $n_i$ but is normalized quadratically). Based on this intuition, we assume low-variance prior weights of the form $p_i(\bx)\propto n_i/s_i$ for $K$Means clustering routers.

For $k$NN routers, since it is difficult to estimate per-query variance $\text{Var}[\epsilon^{(m)}(\bx_\text{tr})]$, we assume the prior probabilities $p_i(\bx)=1/k$ for the $k$ reference elements that are closest to the test query $\bx$, and zero otherwise. 

\subsection{Controlling the Bias-Variance trade-off via Proximity-Based Tilting}

The minimum-variance estimator, denoted by $\bp(\bx)$ in \Cref{subsec:min_var_prior}, suffers from high bias because it does not adequately prioritize reference elements that are close to the test query $\bx$. To mitigate this bias, we introduce an exponentially tilted aggregation that reweights the minimum-variance estimates $\bp(\bx)$ using a proximity penalty $\phi_i(\bx)$, which is a monotonically increasing function of the distance $d(\bx, \br_i)$ to the reference element indexed by $i$. Formally, we transform the low-variance priors $\bp(\bx)$ into new intensities $\theta_i(\bx) \propto p_i(\bx)\,\exp(-\phi_i(\bx)/\tau)$, and normalize to obtain the proximity aware weights $w_i(\bx)=\theta_i(\bx)/\sum_{i \in [|\mathcal{I}|]}\theta_i(\bx)$.

Conceptually, the above proximity-aware weights are the solution to the following optimization problem in \cref{eq:convex_objective_weights}, which trades off between reducing bias (corresponding to the first term) and reducing variance (corresponding to the second term). The proof details are given in \Cref{app:convex_objective_weights}. 
\begin{align}\label{eq:convex_objective_weights}
    &\min_{\bw(\cdot)\in\Delta^{|\mathcal{I}|}} \sum_{i \in [|\mathcal{I}|]} w_i(\bx)\,\phi_i(\bx)+\tau\,D_{\text{KL}}(\bw(\bx)\|\bp(\bx))\\
&\Rightarrow w_i(\bx) \propto p_i(\bx)\,e^{-\phi_i(\bx)/\tau}.\nonumber 
\end{align}

\begin{figure}[t]
    \centering
    \includegraphics[width=0.7\linewidth]{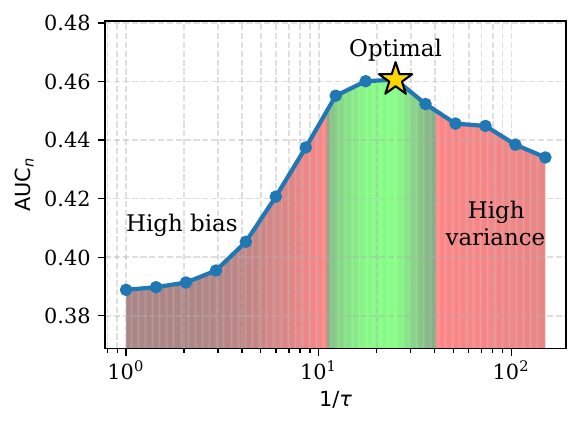}
    \caption{\small Bias-variance tradeoff governed by proximity based prioritization (see \Cref{sec:experiments} for details). Increasing $1/\tau$ strengthens proximity weighting and raises variance (and vice versa for smaller $1/\tau$). Routing performance is measured as area under the mean accuracy-cost curve (AUC), normalized by cost range.
    }
    \label{fig:bias_variance_tradeoff}
\end{figure}

This framework provides a controllable bias–variance tradeoff: increasing $\tau$ drives the weights toward the low-variance priors $\bp(\bx)$, while decreasing $\tau$ emphasizes proximity, thereby reducing bias by preferring semantically similar reference elements. Notably, existing routing schemes emerge as special cases:
\begin{itemize}
    \item $K$Means router with closest cluster assignment is equivalent to $\tau=0$ (no KL penalty).
    \item $k$NN router with uniform averaging over neighbors is equivalent to $\tau\rightarrow\infty$ with uniform priors $\bp(\bx)$ over the $k$ nearest neighbors.
\end{itemize}
\Cref{fig:bias_variance_tradeoff} illustrates bias-variance tradeoff, where varying $\tau$ interpolates between high-bias (small $1/\tau$) and high-variance (large $1/\tau$) regimes, with optimal router performance typically achieved at an intermediate value $1/\tau^*$ that balances proximity confidence against variance reduction. In practice, $\tau$ is tuned on held-out data or adapted to the scale of the distance metric.

Notably, \emph{ProxRouter preserves the original clustering or nearest-neighbors router design}. It is lightweight: the only additional computation involves evaluating low-variance priors and applying the proximity-aware tilt (\Cref{app:delay_overhead}). Moreover, the framework is flexible: it can incorporate a wide range of distance measures in the encoding space, such as Euclidean distance or cosine dissimilarity. Overall, ProxRouter consistently improves routing by (i) lowering variance through principled priors and (ii) reducing bias via proximity-aware weighting.

{\small 
\begin{table*}[tbp]
\centering
\tiny
\setlength{\tabcolsep}{3.0pt}
\caption{Average Accuracies of the models in our considered pool on query datasets used for our experiments. Best two models for each dataset are emphasized in shades of green, the worst two models for that dataset are shown in shades of red. Numbers shown for explicit zero shot prompting (refer \Cref{subapp:prompting_techniques}). 
}
\label{tab:model_performance_grouped}
\begin{tabular}{lccccccccccc}
\toprule

\textbf{\textit{Model \textbackslash Dataset}} & \textbf{SVAMP} & \textbf{GSM8k} & \textbf{HSwag} & \textbf{SciQ} & \textbf{SoQA} & \textbf{WinoG} & \textbf{MedQ} & \textbf{CSQA} & \textbf{LogiQ} & \textbf{BBH} & \textbf{Avg.} \\
\midrule
\textbf{Qwen2-1.5B} & 23.29 & \cellcolor{green!18}68.05 & \cellcolor{red!18}30.60 & 56.15 & 51.05 & 54.30 & \cellcolor{red!35}\textbf{22.10} & 41.95 & 31.20 & \cellcolor{red!35}\textbf{44.80} & 42.35 \\
\textbf{Qwen2-7B} & 51.43 & 29.85 & 77.30 & 90.80 & \cellcolor{green!18}82.05 & 73.10 & 48.50 & \cellcolor{green!18}77.55 & 53.60 & 72.80 & 65.70 \\
\textbf{Deepseek-math-7B} & \cellcolor{green!18}67.86 & \cellcolor{green!35}\textbf{77.05} & 35.20 & 68.05 & 61.35 & 52.40 & 24.20 & 44.30 & 35.70 & 57.60 & 52.37 \\
\textbf{Gemma-2B} & \cellcolor{red!18}22.57 & \cellcolor{red!35}\textbf{4.05} & 31.80 & \cellcolor{red!18}55.50 & \cellcolor{red!18}50.25 & \cellcolor{red!35}\textbf{50.90} & 23.50 & \cellcolor{red!18}36.85 & \cellcolor{red!18}30.00 & 56.80 & \cellcolor{red!18}36.22 \\
\textbf{Gemma-7B} & 52.71 & 13.10 & 50.75 & 81.95 & 70.95 & 57.60 & 32.50 & 62.05 & 38.60 & 64.80 & 52.50 \\
\textbf{Phi-3-small-7B} & 65.43 & 17.55 & \cellcolor{green!18}79.20 & \cellcolor{green!18}94.10 & \cellcolor{green!18}82.05 & \cellcolor{green!35}\textbf{85.30} & \cellcolor{green!18}58.40 & 75.50 & \cellcolor{green!18}54.70 & 76.00 & \cellcolor{green!18}68.82 \\
\textbf{Llama-2-7B} & 39.14 & 7.00 & 47.60 & 76.50 & 63.85 & 51.90 & 26.10 & 49.95 & 38.00 & \cellcolor{red!35}\textbf{44.80} & 44.48 \\
\textbf{Llama-2-13B} & 47.29 & 12.10 & 57.75 & 80.65 & 67.90 & 55.60 & 32.50 & 55.95 & 40.40 & 54.40 & 50.45 \\
\textbf{Llama-3.2-1B} & \cellcolor{red!35}\textbf{14.14} & \cellcolor{red!18}4.50 & \cellcolor{red!35}\textbf{27.35} & \cellcolor{red!35}\textbf{47.40} & \cellcolor{red!35}\textbf{40.60} & \cellcolor{red!18}51.70 & \cellcolor{red!18}22.80 & \cellcolor{red!35}\textbf{25.55} & \cellcolor{red!35}\textbf{26.05} & \cellcolor{red!18}51.20 & \cellcolor{red!35}\textbf{31.13} \\
\textbf{Llama-3.2-3B} & 35.29 & 12.90 & 44.40 & 91.75 & 70.85 & 53.40 & 46.80 & 65.85 & 38.25 & 60.00 & 51.95 \\
\textbf{Llama-3.1-8B} & 37.14 & 44.10 & 53.00 & 91.50 & 72.95 & 58.10 & 46.00 & 68.35 & 43.00 & 64.80 & 57.89 \\
\textbf{Llama-3.3-70B} & \cellcolor{green!35}\textbf{86.29} & 40.80 & \cellcolor{green!35}\textbf{87.05} & \cellcolor{green!35}\textbf{97.65} & \cellcolor{green!35}\textbf{83.40} & \cellcolor{green!18}83.80 & \cellcolor{green!35}\textbf{84.70} & \cellcolor{green!35}\textbf{80.45} & \cellcolor{green!35}\textbf{60.45} & \cellcolor{green!18}79.20 & \cellcolor{green!35}\textbf{78.38} \\
\textbf{Mistral-7B} & 52.00 & 22.55 & 69.55 & 85.55 & 74.50 & 59.30 & 45.40 & 68.15 & 45.85 & 74.40 & 59.72 \\
\textbf{Mixtral-8x7B} & 59.57 & 26.15 & 74.10 & 87.05 & 71.85 & 68.60 & 53.90 & 67.65 & 50.35 & \cellcolor{green!35}\textbf{82.40} & 64.16 \\
\midrule
\textit{Max across models} & 86.29 & 77.05 & 87.05 & 97.65 & 83.40 & 85.30 & 84.70 & 80.45 & 60.45 & 82.40 & 78.38 \\
\textit{Mean across models} & 46.72 & 27.12 & 54.69 & 78.90 & 67.40 & 61.14 & 40.53 & 58.58 & 41.87 & 63.14 & 54.01 \\
\textit{Min across models} & 14.14 & 4.05 & 27.35 & 47.40 & 40.60 & 50.90 & 22.10 & 25.55 & 26.05 & 44.80 & 31.13 \\
\bottomrule
\end{tabular}
\end{table*}
}

\section{Experimental Analysis}
\label{sec:experiments}

Guided by the proximity-aware aggregation described in \Cref{sec:proxrouter}, we design and examine the performance of ProxRouter for both clustering based ($K$Means) and nearest neighbor ($k$NN) settings using corresponding choices of the reference set and its elements. As remarked earlier, a central challenge in router performance is generalization to queries from tasks that are absent or underrepresented in the training query set $\{\bx_\text{tr} : \bx_\text{tr} \in\mathcal{X}_\text{tr}\}$. In the subsequent assessment, we evaluate ProxRouter and corresponding baseline with the same training query set $\mathcal{X}_\text{tr}$, and contextualize the improvement in performance with respect to the full knowledge training query set, modeling $\mathcal{X}_\text{tr} \approx \mathcal{X}$ for an upper bound on routing performance. 

We first describe our experimental setup and evaluation regimes (\Cref{subsec:exp_eval_setup}), and then describe the performance of ProxRouter (in both $K$Means and $k$NN setting) on a variety of tasks (\Cref{subsec:results}) and comment on router retraining.

\subsection{Experimental and Evaluation Setup}\label{subsec:exp_eval_setup}
We conduct our study on ten publicly available language query datasets, choosing multiple inlier/outlier task configurations. Our model pool comprises of 14 diverse LLMs covering a wide spectrum of parameter count and capabilities. A summary of the query datasets and models, along with average accuracies for each model–dataset pair, appears in \Cref{tab:model_performance_grouped}. For query encodings, we employ the MPNet-base sentence encoder \cite{song2020mpnetmaskedpermutedpretraining}. Additional details such as dataset characteristics, LLM attributes, prompting strategies, inference costs, and the sentence-encoder selection are provided in \Cref{app:exp_setup}.

We identify two sources of under representation of queries in the training set $\mathcal{X}_\text{tr}$: Leave-Task-Out scenario where no training queries represent outlier tasks, and Few-Shot-Outlier scenario where very few queries from outlier tasks are present in training set. Leave-Task-Out scenario naturally arises in clustering based routers where no unique cluster represents outlier queries, whereas Few-Shot-Outlier case is typical to nearest neighbor routers where uniform averaging over neighbors obscures information conveyed by few training queries that are closer to outlier test query. 

For clustering based ($K$Means) query routers, we evaluate our approach $K$M-Prox with the closest cluster assignment policy $K$M-Base in the Leave-Task-Out regime. For nearest neighbor based ($k$NN) query routers, we evaluate our approach $k$NN-Prox with the uniform nearest neighbor averaging router $k$NN-Base in the Few-Shot-Outlier scenario. Additionally, we also interpret  improvement in outlier generalization of $K$M-Prox ($k$NN-Prox) with $K$M-AllSee ($k$NN-AllSee) respectively which have seen all queries in training set, modeling the full knowledge case $\mathcal{X}_\text{tr} \approx \mathcal{X}$ and providing an upper bound in routing performance. We also include two router-agnostic baselines: \textit{(i)} Expensive Model routing to the biggest generic model in the pool (Llama 70B in our case), \textit{{(ii)}} Random Routing to any model from the pool.

A preferable router is one which yields higher accuracy values on test queries at the same cost, or alternatively lower costs while maintaining same accuracy level. To quantify router performance based on this observation, we use area under the mean accuracy cost plot normalized by the cost range \cite{hu2024routerbench,jitkrittum2025universalmodelroutingefficient}, a value in $[0,1]$ which we denote by a percentage scale.

\subsection{Results}\label{subsec:results}
\paragraph{$K$M-Prox}
Across all $K$M-Base, $K$M-Prox and $K$M-AllSee, we set number of clusters $K=32$, proximity penalty $\phi_i(\bx) = d(\bx, \mathbf{r}_i)$ as the cosine distance between test query and clusters, and $1/\tau=20$.
 \begin{figure}[!htbp]
    \centering
    \begin{subfigure}[t]{0.49\linewidth}
    \centering
        \includegraphics[width=\linewidth]{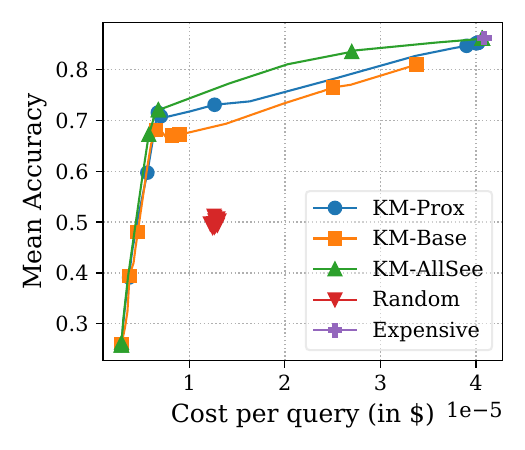}
    \end{subfigure}
    \hfill
    \begin{subfigure}[t]{0.49\linewidth}
    \centering
        \includegraphics[width=\linewidth]{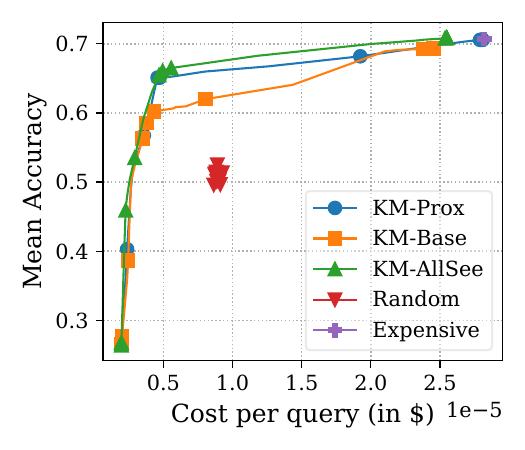}
    \end{subfigure}
    \caption{\small $K$M-Prox, $K$M-Base and $K$M-AllSee router performance on \textit{(left)} MedQA, HellaSwag as outlier tasks and \textit{(right)} LogiQA, CommonSenseQA, BBH-BoolEx as outlier tasks. $K$M-Prox consistently improves routing performance over $K$M-Base without additional training data. 
    }
    \label{fig:kmeans_ood}
\end{figure}

We consider two sets of task distributions for our analysis of ProxRouter on clustering based settings following the Leave-Task-Out generalization, \textit{(i)} Hellaswag and MedQA as outliers and \textit{(ii)} LogiQA, BigBenchHard (BoolEx), CommonsenseQA as outliers. In both cases, outlier tasks are absent in $\mathcal{X}_\text{tr}$, and all other tasks from \Cref{tab:model_performance_grouped} are present as both training queries and test queries. We split the inlier tasks into a 60-40 train-test split, and outlier tasks fully present in the test split. 
\begin{table}[htb]
\centering
\small
\caption{\small Performance (AUC normalized) of $K$M-Prox (ProxRouter) vs. $K$M-Base for chosen sets of outlier tasks. Upper bound denoted by the full knowledge router $K$M-AllSee. ProxRouter improves generalization on outlier queries, with unaffected routing performance on inlier queries across different outlier configurations.}
\label{tab:kmeans_proxrouter}
\setlength{\tabcolsep}{2pt}
\begin{tabular}{@{}ccccc@{}}
\toprule
\multicolumn{1}{c}{\multirow{2}{*}{\shortstack{Outlier Tasks}}} &
\multicolumn{1}{c}{\multirow{2}{*}{Split}} &
\multicolumn{2}{c}{Routing method} &
\multicolumn{1}{c}{Upper Bound} \\
\cmidrule(lr){3-4}\cmidrule(lr){5-5}
 & & $K$M-Base & \cellcolor{green!15}$K$M-Prox & $K$M-AllSee \\
\midrule
\multirow{3}{*}{\shortstack{Hellaswag,\\MedQA}}
  & Outlier & 70.68\% & \cellcolor{green!15}74.88\% & 78.36\% \\
  & Inlier  & 74.62\% & \cellcolor{green!15}74.86\% & 74.63\%\\
  & Overall & 73.04\% & \cellcolor{green!15}75.12\% & 74.87\% \\
\addlinespace
\multirow{3}{*}{\shortstack{LogiQA,\\CSQA, BBH}}
  & Outlier & 63.39\% & \cellcolor{green!15}66.18\% & 67.25\% \\
  & Inlier  & 79.35\% & \cellcolor{green!15}79.92\% & 79.74\% \\
  & Overall & 71.61\% & \cellcolor{green!15}73.46\% & 73.88\% \\
\bottomrule
\end{tabular}
\end{table}

We observe that for both outlier cases (\Cref{fig:kmeans_ood}), $K$M-Prox improves upon the normalized AUC of the mean accuracy cost plot as compared to the $K$M-Base with the same $\mathcal{X}_\text{tr}$ at practically no additional computational costs for routing (\Cref{app:delay_overhead}), and bridges the gap between the $K$M-AllSee full knowledge router upper bound. Notably our approach provides not just overall higher AUC, but provides better accuracy at almost all operating cost regions in the mean accuracy cost plot. Additionally we highlight that all three ($K$M-Base, $K$M-Prox, $K$M-AllSee) perform identically for inlier tasks, indicating unaffected inlier performance of ProxRouter with better generalizability to unseen outlier tasks. Supplementary results in \Cref{app:extended_results}.

\paragraph{$k$NN-Prox}
Across $k$NN-Base, $k$NN-Prox and $k$NN-AllSee, we set number of neighbors $k=100$, proximity penalty $\phi_i(\bx) = d(\bx, \mathbf{r}_i)$ as the cosine distance between test query and its neighbors, and $1/\tau=20$. We consider the Few-shot Outlier case for evaluating ProxRouter in the nearest neighbor based setting where $\mathcal{X}_\text{tr}$ comprises of few ($\sim 25$) queries from math tasks (GSM8k, SVAMP).
\begin{table}[htb]
\centering
\small
\caption{\small Performance (AUC normalized) of $k$NN-Prox (ProxRouter) vs. $k$NN-Base for chosen outlier tasks. Upper bound denoted by the full knowledge router $k$NN-AllSee.}
\label{tab:knn_proxrouter}
\setlength{\tabcolsep}{2pt}
\begin{tabular}{@{}ccccc@{}}
\toprule
\multicolumn{1}{c}{\multirow{2}{*}{\shortstack{Outlier Tasks}}} &
\multicolumn{1}{c}{\multirow{2}{*}{Split}} &
\multicolumn{2}{c}{Routing method} &
\multicolumn{1}{c}{Upper Bound} \\
\cmidrule(lr){3-4}\cmidrule(lr){5-5}
 & & $k$NN-Base & \cellcolor{green!15}$k$NN-Prox & $k$NN-AllSee \\
\midrule
\multirow{3}{*}{\shortstack{GSM8k,\\ SVAMP}}
  & Outlier & 38.55\% & \cellcolor{green!15}46.64\% & 60.77\% \\
  & Inlier  & 77.51\% & \cellcolor{green!15}76.96\% & 79.11\% \\
  & Overall & 63.98\% & \cellcolor{green!15}68.12\% & 74.60\% \\
\bottomrule
\end{tabular}
\end{table}

\begin{figure*}[htbp]
    \centering
    \begin{subfigure}[t]{0.32\textwidth}
        \includegraphics[width=\linewidth]{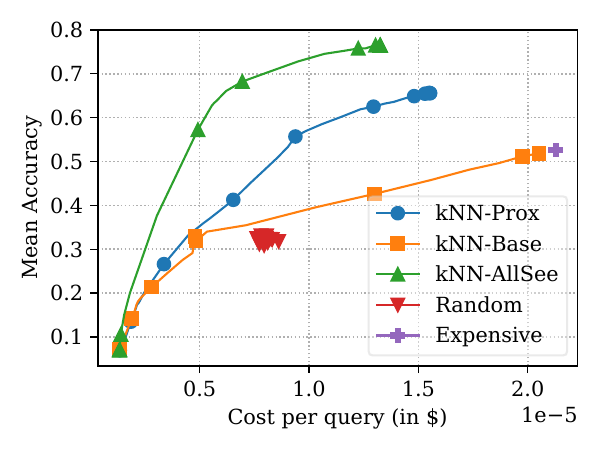}
    \end{subfigure}
    \hfill
    \begin{subfigure}[t]{0.32\textwidth}
        \includegraphics[width=\linewidth]{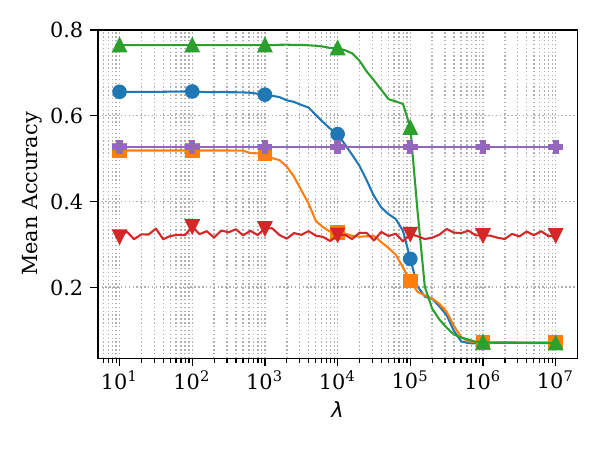}
    \end{subfigure}
    \hfill
    \begin{subfigure}[t]{0.32\textwidth}
        \includegraphics[width=\linewidth]{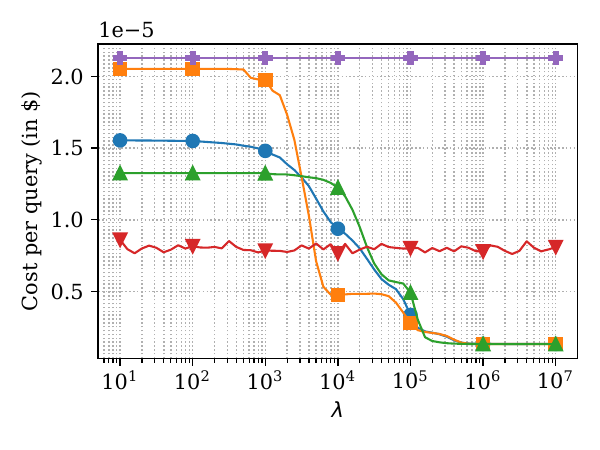}
    \end{subfigure}
    \caption{\small $k$NN-Prox, $k$NN-Base and $k$NN-AllSee router performance on (GSM8k, SVAMP) outliers. \textit{Left} plot describes the mean accuracy-cost plot for outlier queries. \textit{Center} plot describes average accuracy vs $\lambda$ and \textit{right} plot describes average cost per query vs $\lambda$ (\cref{eq:routing_objective}). $k$NN-Prox greatly improves robustness and generalization to outlier queries by proximity-based prioritization of queries in the training set.
    }
    \label{fig:knn_math_ood}
\end{figure*}

\begin{figure}[htb]
    \centering
    \includegraphics[width=0.75\linewidth]{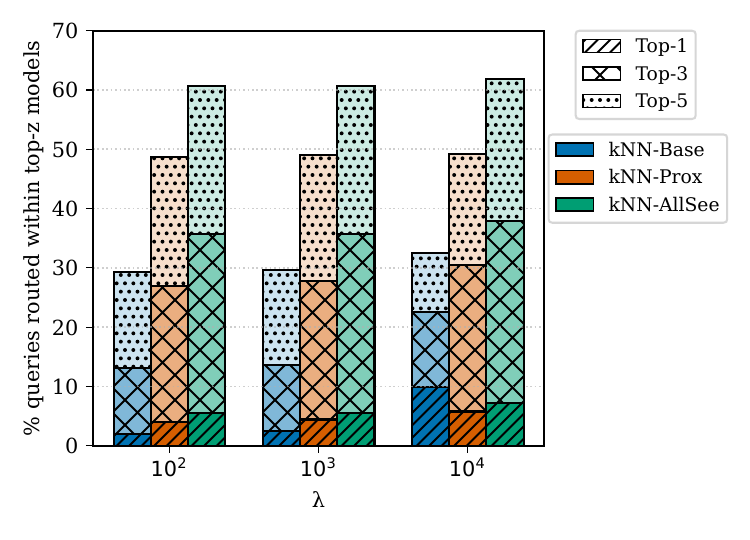}
    \caption{\small Routing Match Accuracies for $k$NN-Prox, $k$NN-Base, $k$NN-AllSee measured against post-hoc top-(1,3,5) most suitable models (by \cref{eq:routing_objective}). Higher is better.}
    \label{fig:querywise_matching_discrete}
\end{figure}

In accordance with previous observations, ProxRouter significantly improves outlier routing performance (\Cref{tab:knn_proxrouter}), increasing the $\text{AUC}_n$ by $8.1$\;pp ($38.5\%$ to $46.6\%$). Remarkably, $k$NN-Prox achieves a higher accuracy value at a significantly lower average cost than $k$NN-Base due to the presence of fine-tuned models that outperform largest model, as depicted in \Cref{fig:knn_math_ood}. \Cref{fig:querywise_matching_discrete} describes the matching accuracy of routing through $k$NN-Base, $k$NN-Prox and $k$NN-AllSee with post-hoc top-$(1,3,5)$ suitable models as per objective in \cref{eq:routing_objective}, indicating that $k$NN-Prox significantly improves routing performance as compared to $k$NN-Base with the same training query set. 

\begin{figure}[htb]
    \centering
    \includegraphics[width=0.8\linewidth]{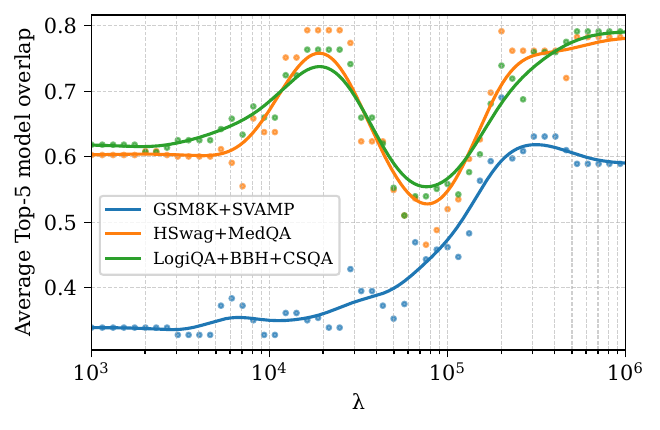}
    \caption{\small Jaccard overlap of top-5 models for three different outlier task sets with corresponding inliers (keeping all but outlier tasks as inliers in the training query set, for each of the three sets). Math tasks (GSM8k, SVAMP) with smaller top-5 model overlap with inlier tasks exhibit larger difference in AllSee and Base router (\Cref{fig:knn_math_ood} and \Cref{tab:knn_proxrouter}).}
    \label{fig:jaccard}
\end{figure}

\paragraph{Outlier gaps and Router Retraining} We observe that the performance gap between the AllSee and the Base router varies with the choice of the outlier task under examination (\Cref{fig:kmeans_ood,fig:knn_math_ood}). These differences may arise because some outlier tasks may demand LLM abilities that co-occur with abilities required for inlier tasks, leading to small gaps as the router already has some idea about handling such outliers, while completely new tasks that require very different LLM skill sets incur larger drops as router has not mapped such LLM abilities in the training query set. To quantify this effect, we measure the similarity of model rankings between inliers and outliers using a simple overlap metric. For any task $t$ and $\lambda$ value, let $S_z(t,\lambda)$ be the set of top-$z$ models by ranked objective value (\cref{eq:routing_objective}). For each outlier task $t_\text{out}$ and inlier task $t_\text{in}$, we define top-$z$ Jaccard similarity 
\begin{equation*}
    J_z(t_\text{out},t_\text{in},\lambda)=\frac{|S_z(t_\text{out},\lambda)\cap S_z(t_\text{in},\lambda)|}{|S_z(t_\text{out},\lambda)\cup S_z(t_\text{in},\lambda)|},
\end{equation*}
and average across all $(t_\text{out},t_\text{in})$ pairs for overall $J_z^\lambda$.

This measure captures why the Base-AllSee gaps differ, as low average $J_z^\lambda$ value indicates that outliers favor very different models than inlier tasks, leading to larger Base-AllSee gaps. Alternatively, higher $J_z^\lambda$ indicates shared model preferences between inliers and outliers, and consequently smaller gaps. Math outliers (GSM8k+SVAMP) have low top-$z$ overlap with inliers (\Cref{fig:jaccard}), whereas the other outlier sets have higher overlap. Higher overlap likely reflects co-existing skills in language models (e.g., reasoning with logical QA, medical with general knowledge); lower overlap reflects a lack of shared skills. This overlap metric also informs \emph{when to retrain} the router, as consistently low average $J_z^\lambda$ for new test queries might be indicative of test-train mismatch of new queries for existing router. Router retraining through evaluation of queries on all models can be triggered when $J_z^\lambda$ falls below a validation chosen threshold for incoming test queries. The choice of threshold can be determined by a cost-aware policy which considers the cost of evaluating outlier queries on all models against the drop in routing performance of existing router. ProxRouter increases robustness between such retraining periods, allowing for a reliable generalization without hindering inlier performance.

\section{Conclusion}
\label{sec:conclusions}
In this work, we address the practical challenge of train–test mismatch in language query routing. We present ProxRouter, a proximity-weighted framework for nonparametric routers that improves robustness to outlier queries while preserving inlier performance. The key idea is to prioritize nearby training queries when estimating each model’s accuracy–cost values for a test query. By exponentially tilting model accuracy-cost estimates, ProxRouter controls the bias-variance trade-off with a single hyperparameter.  We also formalize a broad family of nonparametric language query routers, providing a unified lens for analysis and implementation. Extensive experiments support these claims and underscore ProxRouter’s practical value. We also describe how router retraining decisions can be made by judging the similarity of model rankings on test queries with inlier queries. While this work focuses on nearest neighbours ($k$NN) and clustering ($K$Means) based routers, exploring connections to advanced statistical techniques remains a promising direction.

\section{Acknowledgments}
This work was partially supported by NSF grants CCF 2045694, CNS-2112471, CPS-2111751, ONR grant N00014-23-1-2149, and an AI2C Seed grant. This work used Bridges-2 GPU at the Pittsburgh Supercomputing Center through allocation CIS250429 from the Advanced Cyberinfrastructure Coordination Ecosystem: Services \& Support (ACCESS) program, which is supported by NSF grants \#2138259, \#2138286, \#2138307, \#2137603, and \#2138296 \citep{access}. 


\newpage
\bibliographystyle{plainnat}
\bibliography{References}

\clearpage
\appendix
\thispagestyle{empty}

\onecolumn
\newpage
\aistatstitle{Appendix for "ProxRouter: Proximity-Weighted LLM Query Routing for Improved Robustness to Outliers"}

\appendix

\section{Related Works}\label{sec:related_works}

The language query routing problem is concerned with selecting the appropriate model to generate responses to a query. Alternate thrusts towards efficient usage of large versus small language models additionally focus on intertwining autoregressive generations from large and small models towards the same response, or collecting multiple responses from different models. Speculative decoding \cite{leviathan2023fastinferencetransformersspeculativedecoding} aims to accelerate generation by using smaller models to predict future tokens and larger models to verify them. Multiresponse generation techniques include redundancy exploitation through best-of-n multisampling \cite{ding2025bestrouteadaptivellmrouting}, or through model cascading \cite{dohan2022languagemodelcascades} where response from each model in the sequence determines what model to route to in the next round or to return the response at the current step \cite{dekoninck2025unifiedapproachroutingcascading, chen2023frugalgptuselargelanguage}. Preference aligned routers match human rating of generated responses from different models \cite{ong2025routellmlearningroutellms, frick2025prompttoleaderboard} evaluated on ChatbotArena \cite{chiang2024chatbotarenaopenplatform}.  \cite{ding2024hybridllmcostefficientqualityaware,ong2025routellmlearningroutellms} consider routing across two language models, providing fundamental insights extendable to multiple model setting. Bandit and decision making informed router design initiates dynamic inference \cite{li2025llmbanditcostefficientllm} \cite{treacle_thrifty_reasoning} \cite{poon2025onlinemultillmselectioncontextual_meta_llm}, improving performance on queries under budget constraints. Graphical models over LLMs and their outputs represent contextual information and present a weak supervised approach \cite{guha2024smoothielabelfreelanguage,feng2025graphroutergraphbasedrouterllm}. \cite{jitkrittum2025universalmodelroutingefficient} accentuates the adaptability to a dynamic LLM pool, further highlighted by \cite{zhang2025avengerssimplerecipeuniting} through model capability profiling. Model ability comparison through Elo scores \cite{zhao2024eagleefficienttrainingfreerouter} and Item Response Theory \cite{song2025irtroutereffectiveinterpretablemultillm} presents an interpretable approach, furthering the understanding of router behavior. Other techniques involve utilizing query tagging among tasks \cite{chen2025tagrouterlearningroutellms}, distilling reward models \cite{lu2023routingexpertefficientrewardguided_zooter}, POMDP modeling \cite{aggarwal2025automixautomaticallymixinglanguage}, introducing contrastive loss \cite{chen2024routerdcquerybasedrouterdual} etc.

\paragraph{Parametric and Nonparametric Routers} Structurally, most query routing approaches utilize fixed dimensional query encodings generated through deterministic encoder models \cite{cer2018universalsentenceencoder}, \cite{paraphrase_albert_reimers-2019-sentence-bert} which forms the backbone shared by both learned (parametric) routers and learning-free (nonparametric) routers. Building above these encodings, parametric approaches learn small multi-layer perceptrons to predict model correctness or cost \cite{hu2024routerbench,ding2024hybridllmcostefficientqualityaware, ding2025bestrouteadaptivellmrouting, zhuang2024embedllmlearningcompactrepresentations, sakota_2024_flyswat,huang2025routerevalcomprehensivebenchmarkrouting}. Additionally, parametric approaches also examine Bradley Terry ranking, matrix factorization and multihead classifiers \cite{ong2025routellmlearningroutellms}, or train smaller language model for reward modeling, task identification or hardness determination \cite{treacle_thrifty_reasoning, chen2025tagrouterlearningroutellms}. Nonparametric routers work directly on the embedding space of training queries without learning additional parameters for making routing decisions. To canonical types are prevalent: clustering based (eg. $K$Means) routers \cite{zhang2025avengerssimplerecipeuniting, avengers_pro_zhang2025gpt5makingllmscheaper, jitkrittum2025universal, hu2024routerbench} and nearest neighbor based (eg. $k$NN) routers \cite{li2025rethinkingpredictivemodelingllm, zhuang2024embedllmlearningcompactrepresentations, jitkrittum2025universalmodelroutingefficient, stripelis2024tensoroperaroutermultimodelrouter}.  Empirical findings describe nonparametric router performance as almost equal or even better than parametric counterparts. The surprising competitiveness of simple approaches likely stems from the richness of query encoding representations generated through encoder models (\cref{fig:encoding_queries}), which greatly simplifies query to model ability mapping. Furthermore, clustering and nearest neighbor routers are incremental by design, signifying that newer queries and language models can be easily added to the design without having to retrain the router as in parametric approaches. For these reasons, we adopt and focus on the nonparametric router paradigm in this work.

\paragraph{Robust Routing} A unifying observation across multiple techniques is that query routing hinges on accurate predictions of language model correctness across a range of queries. An ideal router makes accurate routing decisions across models on all types of queries, but such a router is infeasible in practice (\cref{sec:introduction}). Prior works often quantify outlier performance drop with their approaches, but seldom aim to explicitly improve robustness and generalizability to queries from unseen tasks. We build on these observations and use theoretical insights to unify the analysis of a broad category of nonparametric routers, and utilize proximity aware aggregation coupled with variance reduction to generalize routing. We also underscore the scope of our work as being designing robust model accuracy and cost estimators, and our work can be readily plugged into multiple strategies such as model cascading, best-of-n, dynamic routing etc. 

\paragraph{Scope and Methodological Focus} A broad toolbox of existing statistical approaches can be used to model the problem of language query routing, including but not limited to fuzzy/soft $K$Means \cite{FERRARO2024110_fuzzy_kmeans}, spectral clustering \cite{NIPS2001_801272ee_spectral_clustering}, kernel smoothing \cite{WandJones1995_kernel_smoothing}, mode seeking methods \cite{400568_mode_seeking}, Bayesian estimators \cite{Berger1985_bayesian} etc. These techniques have not been systematically explored for language query routing or widely adopted to our knowledge. Accordingly, we deliberately focus our study on simplistic $K$Means \cite{IKOTUN2023178_kmeans} and $k$NN \cite{60c19788-1128-3b5f-9275-2d63cc155832_knn_orig} based query routers as they are dominantly used across query routing literature and offer simplicity and interpretability. We contribute by building on these primitive yet well performing and well studied algorithms by improving their robustness and generalizability. Our unified representation for non parametric routers (\cref{subsec:general_form}) is compatible with most of the aforementioned statistical techniques, and we highlight their examination as a promising future direction.

\section{Experimental Setup} \label{app:exp_setup}
Our query datasets evaluate language model abilities across multiple categories:  \textbf{CommonSenseQA} \cite{talmor-etal-2019-commonsenseqa}, \textbf{GSM8K} \cite{cobbe2021gsm8k}, \textbf{Hellaswag} \cite{zellers-etal-2019-hellaswag}, \textbf{LogiQA}\cite{liu2020logiqachallengedatasetmachine}, \textbf{MedQA} \cite{medqa_jin2021disease}, \textbf{SciQ} \cite{sciq_welbl2017crowdsourcingmultiplechoicescience}, \textbf{Social\_i\_QA} \cite{sap2019socialiqacommonsensereasoningsocial}, \textbf{SVAMP} \cite{svamp_patel-etal-2021-nlp}, \textbf{Winogrande} \cite{winogrande_sakaguchi2019winograndeadversarialwinogradschema}, \textbf{BBH} (BoolEx) \cite{suzgun2022challengingbigbenchtaskschainofthought}. Detailed prompting techniques are elaborated later in the section.
Each dataset comprises of hundreds to thousands of queries. To ensure a balanced comparison across tasks, we select a subset of each for training and evaluation, as described in the \Cref{subapp:dataset_split}. 
We intentionally exclude multi-subtask datasets such as ARC-Challenge \cite{arc-challenge_clark2018thinksolvedquestionanswering} and MMLU \cite{mmlu_hendrycks2021measuringmassivemultitasklanguage} from our analysis in order to benchmark generalizability to unseen tasks. Including such datasets risks inadvertent data leakage, since they encompass queries from numerous subtasks (e.g., MMLU includes four distinct math-related subtasks among its 57), potentially overlapping with the test distribution and compromising evaluation integrity.\\

We choose language models across different capability ranges as indicated by their number of parameters: \textit{(i)} \textbf{<3B} Params: Llama-3.2-1B-Instruct \cite{grattafiori2024llama3herdmodels}, Qwen2-1.5B-Instruct \cite{yang2024qwen2technicalreport}, Gemma-2b-it \cite{gemmateam2024gemmaopenmodelsbased}, Llama-3.2-3B-Instruct \cite{grattafiori2024llama3herdmodels}
    \textit{(ii)} \textbf{7B} Params: Phi-3-small-8k-instruct \cite{abdin2024phi3technicalreporthighly}, Qwen2-7B-Instruct \cite{yang2024qwen2technicalreport}, Deepseek-math-7b-instruct \cite{shao2024deepseekmathpushinglimitsmathematical}, Llama-2-7b-chat-hf \cite{touvron2023llama2openfoundation}, Gemma-7b-it \cite{gemmateam2024gemmaopenmodelsbased}, Mistral-7B-Instruct-v0.3 \cite{jiang2023mistral7b}, Llama-3.1-8B-Instruct \cite{grattafiori2024llama3herdmodels} 
    \textit{(iii)} \textbf{13B} Params: Llama-2-13b-chat-hf \cite{touvron2023llama2openfoundation}
    \textit{(iv)} \textbf{>50B} Params: Mixtral-8x7B-Instruct-v0.1 \cite{jiang2024mixtralexperts}, Llama-3.3-70B-Instruct \cite{grattafiori2024llama3herdmodels}. 
Our LLM ensemble consists of standard autoregressive models, chat and instruct fine-tuned models, mixture-of-experts based models and task-specific finetuned models. The inference costs for chosen language models is provided in \Cref{subapp:price_per_token} in $\$$/M tokens.\\ 

Our analysis considers query embeddings generated from the following language encoder models: MiniLM L6 \cite{wang2020minilmdeepselfattentiondistillation}, Paraphrase Albert small \cite{paraphrase_albert_reimers-2019-sentence-bert}, MPNet base \cite{song2020mpnetmaskedpermutedpretraining}, DistillRoberta \cite{distill_roberta_Sanh2019DistilBERTAD} with fixed embedding dimensions being 384, 768, 768, 768 respectively. We observe that router performance is relatively unaffected by the choice of the underlying sentence encoder model (\Cref{tab:enc_model_routerperf}). For the sake of uniformity, we utilize MPNet Base for our experiments.

{\begin{table}[htbp]
\centering
\small
\caption{\small Base Router performance by encoder model (normalized AUC for mean acc-cost plot) $K=32$ for $K$Means router, nearest neighbours $k=100$ for $k$NN router. Notably, routing performance is relatively unnaffected by the choice of encoder model. We utilitize MPNet Base encoder throughout our experiments.}
\label{tab:enc_model_routerperf}
\begin{tabular}{@{}l r r cc@{}}
\toprule
\textbf{Encoder} & \textbf{Enc Dim} & \textbf{$K$Means} & \textbf{$k$NN} \\
\midrule
MiniLM L6          & 384  & 76.3\% & 75.7\% \\
Para Albert Small  & 768  & 77.1\% &  75.9\% \\
MPNet Base         & 768  & 76.7\% & 75.9\% \\
DistillRoberta     & 768  & 77.2\% & 75.7\% \\
\bottomrule
\end{tabular}
\end{table}
}

\subsection{Pricing Structure for Language Models}\label{subapp:price_per_token}
\Cref{tab:token-prices-per-million} denotes the pricing structure of utilized language models in terms of unit input and output token prices \cite{artificialanalysisleaderboard}.
\begin{table}[t]
  \centering
  \small
  \caption{Inference pricing in USD per 1M tokens.}
  \label{tab:token-prices-per-million}
  \begin{tabular}{lcc}
    \toprule
    \textbf{Model} & \textbf{Input (\$/1M tok)} & \textbf{Output (\$/1M tok)} \\
    \midrule
    Qwen/Qwen2-1.5B-Instruct             & 0.015 & 0.030 \\
    Qwen/Qwen2-7B-Instruct               & 0.040 & 0.080 \\
    deepseek-ai/deepseek-math-7b-instruct& 0.040 & 0.080 \\
    google/gemma-2b-it                   & 0.015 & 0.030 \\
    google/gemma-7b-it                   & 0.030 & 0.050 \\
    meta-llama/Llama-2-13b-chat-hf       & 0.050 & 0.100 \\
    meta-llama/Llama-2-7b-chat-hf        & 0.020 & 0.040 \\
    meta-llama/Llama-3.1-8B-Instruct     & 0.030 & 0.050 \\
    meta-llama/Llama-3.2-1B-Instruct     & 0.010 & 0.020 \\
    meta-llama/Llama-3.2-3B-Instruct     & 0.015 & 0.030 \\
    meta-llama/Llama-3.3-70B-Instruct    & 0.150 & 0.300 \\
    microsoft/Phi-3-small-8k-instruct    & 0.025 & 0.050 \\
    mistralai/Mistral-7B-Instruct-v0.3   & 0.020 & 0.050 \\
    mistralai/Mixtral-8x7B-Instruct-v0.1 & 0.100 & 0.250 \\
    \bottomrule
  \end{tabular}
\end{table}
\subsection{Prompting Techniques}\label{subapp:prompting_techniques}

We utilize zero shot prompting technique for queries from all tasks, including math queries which elicit generated response with final numeric answer in the output sequence and multiple choice questions which only require letter options corresponding to the correct answer. All models share the same evaluation approach for a given task, depending on the type and cardinality of the set of responses. We host and evaluate all opensource models locally on an array of four NVIDIA Tesla V100s (32GB) \cite{nvidia-tesla-v100-datasheet-2018} and four H100s (80GB) \cite{nvidia-h100-pcie-product-brief-2022}. We use vLLM inference machine \cite{kwon2023efficientmemorymanagementlarge_vllm} for generating responses for provided prompts for the language models, with default suggested LLM hyperparameter settings and chat templates wherever applicable.

\subsubsection{GSM8K}

\begin{promptbox}
\small 
Solve the following math problem and provide a numerical answer.\\
\textbf{Question:} Jen and Tyler are gymnasts practicing flips. Jen is practicing the triple-flip while Tyler is practicing the double-flip. Jen did sixteen triple-flips during practice. Tyler flipped in the air half the number of times Jen did. How many double-flips did Tyler do?\\
Give only the final answer without explanation.\\
\textbf{Answer:}
\end{promptbox}

\subsubsection{HellaSwag}
\begin{promptbox}
\small
Find the most likely ending of the given question.\\
\textbf{Question:} A woman is wearing a white robe and a black belt. She does karate moves in her room. she\\
\textbf{Choices:}
\begin{adjustwidth}{2em}{0pt}
Option (A): gets into a cage in the middle of her room.\\
Option (B): kicks her legs up several times.\\
Option (C): throws an object off in the distance.\\
Option (D): watches tv and eats ice cream on the couch.
\end{adjustwidth}
Return the correct option letter in parenthesis () without any explanation:
\end{promptbox}

\subsubsection{MedQA}
\begin{promptbox}
\small
Please answer with the correct option letter.\\
\textbf{Question:} A 46-year-old man comes to the physician because of a 2-month history of hoarseness and drooling. Initially, he had difficulty swallowing solid food, but now he has difficulty swallowing foods like oatmeal as well. During this period, he also developed weakness in both arms and has had an 8.2 kg (18 lb) weight loss. He appears ill. His vital signs are within normal limits. Examination shows tongue atrophy and pooled oral secretions. There is diffuse muscle atrophy in all extremities. Deep tendon reflexes are 3+ in all extremities. Sensation to pinprick, light touch, and vibration is intact. An esophagogastroduodenoscopy shows no abnormalities. Which of the following is the most likely cause of this patient's symptoms?\\
\textbf{Choices:}
\begin{adjustwidth}{2em}{0pt}
(A) Multiple cerebral infarctions\\
(B) Autoimmune destruction of acetylcholine receptors\\
(C) Demyelination of peripheral nerves\\
(D) Destruction of upper and lower motor neurons\\
(E) Dilation of the central spinal canal
\end{adjustwidth}
Return the answer option letter in parenthesis () without any explanation.\\
\textbf{Answer:}
\end{promptbox}

\subsubsection{SciQ}
\begin{promptbox}
\small
Please answer with the correct option letter.\\
\textbf{Question:} Nearly all protists exist in some type of aquatic environment, including freshwater and marine environments, damp soil, and even snow. Several protist species are parasites that infect animals or plants. A few protist species live on dead organisms or their wastes, and contribute to what?\\
\textbf{Choices:}
\begin{adjustwidth}{2em}{0pt}
(A) greenhouse gas\\
(B) habitat loss\\
(C) spontaneous mutation\\
(D) their decay
\end{adjustwidth}
Return only the correct option letter in parenthesis () without any explanation.\\
\textbf{Answer:}
\end{promptbox}

\subsubsection{Social iQA}
\begin{promptbox}
\small
Please answer with the correct option letter based on the given context.\\
\textbf{Context:} Skylar met their requirements and ended up getting hired for the job.\\
\textbf{Question:} How would you describe Skylar?\\
\textbf{Choices:}
\begin{adjustwidth}{2em}{0pt}
(A) excited to start\\
(B) as someone that got the job\\
(C) proud of herself
\end{adjustwidth}
Return only the correct option letter in parenthesis () without any explanation.\\
\textbf{Answer:}
\end{promptbox}

\subsubsection{SVAMP}
\begin{promptbox}
\small
Solve the following math problem and provide a numerical answer. \\
\textbf{Question:} Jack received 6 emails and 8 letters in the morning. He then received 2 emails and 7 letters in the afternoon. How many more letters did Jack receive in the morning than in the afternoon?. \\
Give only the correct answer without explanation.\\
\textbf{Answer:}
\end{promptbox}

\subsubsection{Winogrande}
\begin{promptbox}
\small
Choose the correct option letter that fills in the underscore blank in the question.\\
\textbf{Question:} The tailor called Megan to say her coat was finished, so she asked her assistant Patricia to get it. \_ drove the car to the tailor shop.\\
\textbf{Choices:}
\begin{adjustwidth}{2em}{0pt}
(A) Megan\\
(B) Patricia
\end{adjustwidth}
Return only the letter corresponding to the correct option in parenthesis () without any explanation.\\
\textbf{Answer:}
\end{promptbox}

\subsubsection{Commonsense QA}
\begin{promptbox}
\small
Please answer with the correct option letter.\\
\textbf{Question:} John looked for shade but couldn't find any. Where might he be?\\
\textbf{Choices:}
\begin{adjustwidth}{2em}{0pt}
(A) sunny place\\
(B) summer\\
(C) direct sunlight\\
(D) bright sunshine\\
(E) full sunlight
\end{adjustwidth}
Return only the correct option letter in parenthesis () without any explanation.\\
\textbf{Answer:}
\end{promptbox}

\subsubsection{LogiQA}
\begin{promptbox}
\small
Please answer with the correct option letter based on the given context.\\
\textbf{Context:} A and B walk from the library to the classroom at the same time. A walks halfway and runs halfway. B walks halfway and runs halfway. If both walk, the speed is the same.\\
\textbf{Question:} then?\\
\textbf{Choices:}
\begin{adjustwidth}{2em}{0pt}
(A) A arrives in the classroom first.\\
(B) B arrives in the classroom first.\\
(C) A and B arrive at the classroom at the same time.\\
(D) Unable to judge.
\end{adjustwidth}
Return only the correct option letter in parenthesis () without any explanation.\\
\textbf{Answer:}
\end{promptbox}

\subsubsection{Big Bench Hard (Boolean Expressions)}
\begin{promptbox}
\small
Choose the correct option letter corresponding to the boolean value of the question.\\
\textbf{Question:} not ( True ) and ( True ) is\\
\textbf{Choices:}
\begin{adjustwidth}{2em}{0pt}
(A) True\\
(B) False
\end{adjustwidth}
Return only the correct option letter in parenthesis () without any explanation.\\
\textbf{Answer:}
\end{promptbox}

\subsection{Taskwise Query Composition and Structure}\label{subapp:dataset_split}
Refer \Cref{tab:dataset-attrs} for details.

\begin{table}[!htbp]
\centering
\scriptsize
\setlength{\tabcolsep}{4pt} 
\caption{\small Dataset attributes used in our evaluation. Bracketed  \texttt{answer\_type} values denotes space of correct answers (number of options, or symbolically $\infty$ for numerical responses).}
\label{tab:dataset-attrs}
\begin{tabular}{@{}l*{10}{c}@{}}
\toprule
\textbf{Attribute} &
\textbf{SVAMP} & \textbf{GSM8k} & \textbf{HSwag} & \textbf{SciQ} & \textbf{SoQA} &
\textbf{WinoG} & \textbf{MedQ} & \textbf{CSQA} & \textbf{LogiQ} & \textbf{BBH} \\
\midrule
\texttt{num\_queries} & 700 & 2000 & 2000 & 2000 & 2000 & 1000 & 1000 & 2000 & 2000 & 125 \\
\texttt{answer type} & num($\infty$) & num($\infty$) & mcq(4) & mcq(4) & mcq(3) & mcq(2) & mcq(5) & mcq(5) & mcq(4) & mcq(2) \\
\bottomrule
\end{tabular}
\end{table}

\section{ProxRouter Computational Overhead Analysis}\label{app:delay_overhead}
We measure the delay overhead for routing a single query through ProxRouter across three different scenarios. We analyze \textit{(a)} the delay for nearest neighbor based scenario with varying number of training queries, \textit{(b)} the delay for nearest neighbor based routing across different choices of $k$, \textit{(c)} delay for clustering based router for varying number of clusters; all for multiple choice for encoding dimension $d_\text{enc}\in\{128,256,384,512,768\}$. $k$NN-Prox computational graph includes fetching neighbors--calculating tilted aggregation weights--estimating model utilities--calculating argmax over models. $K$M-Prox computational graph involves calculating the cluster weights--computing the estimates of model objective and finally taking an argmax. 
\begin{figure}[htbp]
  \centering
  \begin{subfigure}[t]{0.32\textwidth}
    \centering
    \includegraphics[width=\linewidth]{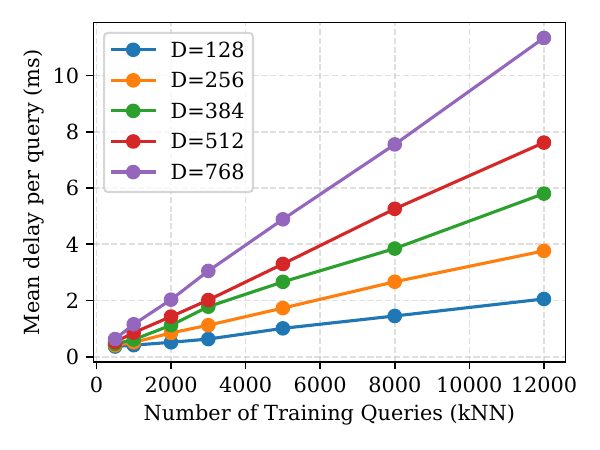}
  \end{subfigure}\hfill
  \begin{subfigure}[t]{0.32\textwidth}
    \centering
    \includegraphics[width=\linewidth]{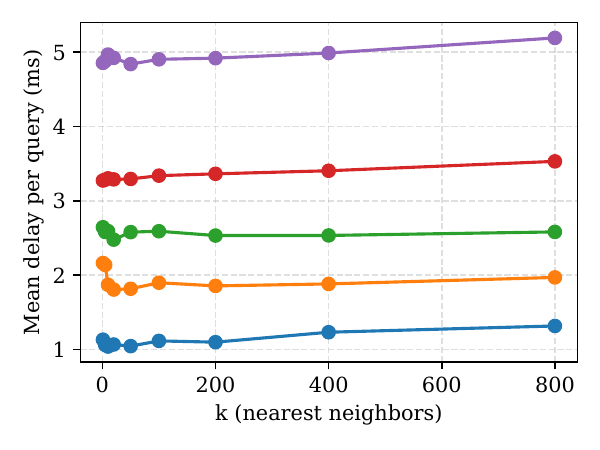}
  \end{subfigure}\hfill
  \begin{subfigure}[t]{0.32\textwidth}
    \centering
    \includegraphics[width=\linewidth]{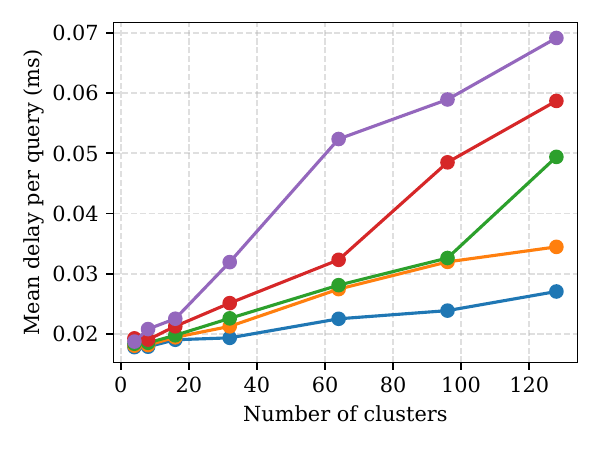}
  \end{subfigure}
  \caption{Routing overhead for ProxRouter (in milliseconds). \textit{Left} plot describes delay of $k$NN-Prox for varying number of training queries $|\mathcal{X}_\text{tr}|$, \textit{center} plot describes $k$NN-Prox latency for varying number of nearest neighbors $k$, \textit{right} plot depicts $K$M-Prox latency for different number of clusters. In all cases, the overhead is roughly 100x smaller than the LLM generation latency, which is in the order of seconds due to autoregressive nature of generation \cite{kwon2023efficientmemorymanagementlarge_vllm}.}
  \label{fig:delay_overhead}
\end{figure}

We utilize FAISS \cite{douze2025faisslibrary} and Scikit-Learn \cite{scikit-learn} for index search over query points in the encoding space. For stable comparison across different parameters values, we pin all math libraries to one CPU thread which avoids multithreading jitters and dynamic thread allocation delays. 

We highlight that the time delay overhead is in the millisecond range, and thus two orders of magnitude smaller than the usual generation latency of LLMs due to their autoregressive nature. Additionally, we only need to store the training query encodings and model accuracy-cost values, which has a memory footprint of $35\text{MB}$ for float32 representation of $768$-dimensional encodings for $10,000$ training queries. ProxRouter computational graph can easily fit on a single CPU thread, further underscoring its efficiency. Hence, ProxRouter practically adds negligible computational and time overhead to query routing.

\section{Minimum Variance Priors} \label{app:min_var_prior}
For a general set of samples $\{ x_i=(\mu + \varepsilon_i): i \in [n] \}$, with zero mean noise $\text{Var}[\varepsilon_i] = \sigma_i^2$, any unbiased estimate or weighted average from the family
\begin{equation*}
    \widehat{\mu} = \sum_{i\in [n]} w_i \;x_i \;, \qquad w_i\geq 0, \;\;\sum_i w_i = 1.
\end{equation*}
has variance $\sum_i w_i^2 \;\sigma_i^2$. For the uniform sample variance case ($\sigma_i^2 = \sigma_j^2 \;\; \forall \;i,j \in [n]$), the variance of the estimate $\widehat{\mu}$ is 
\begin{equation*}
    \text{Var}(\widehat{\mu}) = \sum w_i^2 \sigma_i^2
\end{equation*}
Any deviation from uniform averaging weights over the same $[n]$ points yields a strictly higher variance estimate for $\widehat{\mu}$.\\

For heteroscedastic noise case, optimal least variance estimate 
weights $w_i \propto 1/\sigma_i^2$, and the minimum variance estimator is 
\begin{equation*}
    \widehat{\mu} = \sum_{i \in [n]} w_i x_i = \sum_i \frac{1/\sigma_i^2}{\sum 1/\sigma_i^2} x_i
\end{equation*}
$\text{Var}(\widehat{\mu}) = 1/ \sum_i \sigma_i^{-2}$. Any deviation from these aggregation weights in the heteroscedastic setting yields a strictly higher variance estimate. The above result can be shown by using the Lagrange multipliers method for minimizing total variance $\sum_i w^2_i \sigma_i^2$ subject to unbiasedness constraint $\sum_i w_i =1$, and finding the stationary point for
\begin{equation*}
    \mathcal{L}(\mathbf{w}, \lambda) = \sum_i w_i^2 \sigma_i^2 + \lambda \big(  \sum_i w_i -1\big)
\end{equation*}
which results in 
\begin{equation*}
    \frac{\partial \mathcal{L}}{\partial w_i} = 2 w_i \sigma_i^2 + \lambda = 0 \quad\Rightarrow\quad w_i = - \frac{\lambda}{2\sigma_i^2}
\end{equation*}
And normalizing weights provides $w_i = \sigma_i^{-2} / \sum_i \sigma_i^{-2}$.

Thus, our proximity-weighted aggregation estimator suffers from the unavoidable increase in variance in exchange for a smaller bias in the estimate. Crucially, we provide a tunable knob through the $\tau$ parameter that controls such bias-variance tradeoff and tames the squared error in objective value estimation.

\section{Proof of \cref{eq:convex_objective_weights}}\label{app:convex_objective_weights}
\paragraph{To Prove:} \begin{align*}
    \min_{\bw(\cdot)\in\Delta^{|\mathcal{I}|}} &\sum_{i\in \{1,\cdots, |\mathcal{I}|\}} w_i(\bx)\,\phi_i(\bx)+\tau\,D_{\text{KL}}(\bw(\bx)\|\bp(\bx))\\
\quad&\Rightarrow w_i(\bx) \propto p_i(\bx)\,e^{-\phi_i(\bx)/\tau}.\nonumber 
\end{align*}
\paragraph{Proof:}
The objective is strictly convex on $\Delta^{|\mathcal{I}|}$, so the minimizer is unique.
Form the Lagrangian with the equality constraint $\sum_i w_i=1$:
\[
\mathcal L(\bw,\lambda)
= \sum_i w_i\phi_i + \tau\sum_i w_i\log\frac{w_i}{p_i}
+ \lambda\left(\sum_i w_i-1\right).
\]
Stationarity ($\partial\mathcal L/\partial w_i=0$) gives, for any $i$ with $w_i>0$,
\[
\phi_i + \tau\Big(1+\log\tfrac{w_i}{p_i}\Big)+\lambda=0
\quad\Longrightarrow\quad
w_i \;=\; C\,p_i\,e^{-\phi_i/\tau},\ \ C:=e^{-(\lambda+\tau)/\tau}.
\]
Enforcing $\sum_i w_i=1$ yields
$C=\big(\sum_j p_j e^{-\phi_j/\tau}\big)^{-1}$ and hence
\[
w_i^*=\frac{p_i\,e^{-\phi_i/\tau}}{\sum_j p_j\,e^{-\phi_j/\tau}}.
\] \qed

\section{Bias Variance Decomposition}\label{app:bias_variance_tradeoff}

Consider the objective function in eq \eqref{eq:routing_objective} to be calculated/estimated for any query $\bx\in\mathcal{X}$, model $m$ 
\begin{equation*}
    U^{(m)}(\bx) = \text{acc}^{(m)}(\bx) - \lambda \,\text{cost}^{(m)}(\bx)
\end{equation*}
As the objective value is a random quantity due to the inherent uncertainty in generation from language models, leading to variable response correctness and output lengths (and consequently costs), we decompose the objective into 
\begin{equation*}
    U^{(m)}(\bx) = \mathbb{E}[ U^{(m)}(\bx)] + \epsilon^{(m)}(\bx) 
\end{equation*}
where $\overline{U}^{(m)} (\bx) = \mathbb{E}[ U^{(m)}(\bx)]$ is the conditional expectation of the objective for model $m$ given query $\bx$, and $\epsilon^{(m)}(\bx)$ is the zero mean noise component of each objective value associated to the randomness in generation. 

\subsection{Assumptions}

\begin{assumption} \label{assumption:smooth_objective} (\textit{Local Smoothness Under Dissimilarity Measure}) For any given queries $\bx_1$ and $\bx_2$, the true objective value difference is upper bounded by a global constant times the distance/dissimilarity between the queries. 
\begin{equation}
    |  U^{(m)}(\bx_1) -  U^{(m)}(\bx_2) | \leq L \; d(\bx_1
,\bx_2) \quad \forall\; \bx_1,\bx_2\in \mathcal{X}; \;\; m\in \mathcal{M}
\end{equation}
Where $d(\cdot, \cdot)$ is any measure that captures dissimilarity between queries in the encoding space. In our analysis, we concern ourselves with dissimilarity measures in the encoding space based on the training objectives of the encoder models (such as cosine distance, Euclidean distance etc.). 
\end{assumption}

\begin{assumption}\label{assumption:unbiased_centroid} (\textit{Centroids as unbiased Cluster Summaries}) For any cluster $c_i$ for $i\in [|\mathcal{I}|]$ of query embeddings, the average of objective functions at query points in a cluster acts as an unbiased estimator of the objective value at the centroid. 
\begin{equation*}
    \mathbb{E} \Big[ \frac{1}{n_i}\sum_{\bx_\text{tr} \in c_i} U^{(m)}(\bx_\text{tr})\Big] \triangleq  \mathbb{E} \left[V_i^{(m)}\right] = \overline{U}^{(m)}(\br_i)
\end{equation*}
where $\br_i$ is the cluster centroid for cluster $c_i$ and $V^{(m)}_i$ is the average objective value for queries in a cluster as per the reference set notation.
\end{assumption}
\begin{assumption}\label{assumption:independent_noise} (\textit{Independence of Evaluation Noise $\epsilon^{(m)}(\bx)$}) The noise associated with the objective value is independent for all queries, training and test alike.
\end{assumption}

\begin{assumption} \label{assumption:fixed_design} (\textit{Fixed Design Setting}) Router is built using a fixed corpus of training queries, with deterministic encodings and observed correctness values and incurred costs.
\end{assumption}

\subsection{General Form of Non-Parametric Routers}

Utilizing the reference set notation, we denote the family of nonparametric routers as estimators of model objective values for any test query $\bx$ as
\begin{equation}
    \widehat{U}^{(m)}(\bx) = \sum_{i \in |\mathcal{I}|} w_i(\bx) V^{(m)}_i 
\end{equation}
where $w_i(\bx)\geq 0, \;\;\sum_{i \in |\mathcal{I}|} w_i(\bx) = 1$. Details elaborated in \cref{subsec:general_form}.

\subsection{Explicit Bias Variance Tradeoff}

\paragraph{Objective:} Find a nonparametric estimator $\widehat{U}^{(m)}(\bx)$ from the above family of estimators that approximates the expected objective value $\overline{U}^{(m)}(\bx)$ for any test query $\bx\in\mathcal{X}$, given the observed $U^{(m)}(\bx_\text{tr})$ values for training queries $\bx_\text{tr}\in\mathcal{X}_\text{tr}$ for all models $m\in\mathcal{M}$. The final routing decision is based on these estimates as $m^*(\bx) \leftarrow \argmax_m \widehat{U}^{(m)}(\bx)$.
\paragraph{Squared Error Decomposition:} We represent the mean squared error loss between the estimate $\widehat{U}^{(m)}(\bx)$ and actual value $U^{(m)}(\bx)$ for a given test query $\bx\in\mathcal{X}$ as 
\begin{align}\label{eq:squared_error}
    \text{Error} (\widehat{U}^{(m)}(\bx), U^{(m)}(\bx)) &= \mathbb{E}\big[  \big(\widehat{U}^{(m)}(\bx) -  U^{(m)}(\bx)\big)^2  \big] \\
    &= \mathbb{E}\big[ \big(\widehat{U}^{(m)}(\bx) - (\overline{U}^{(m)}(\bx) + \epsilon^{(m)}(\bx)) \big)^2 \big] \nonumber\\
    &= \mathbb{E}\big[ \big(\widehat{U}^{(m)}(\bx) - \overline{U}^{(m)}(\bx) \big)^2\big] + \mathbb{E}[\big(\epsilon^{(m)}(\bx)\big)^2] \nonumber
\end{align}
where the expectation is over the randomness in generation over the text query $\bx$ and observations of model accuracy and cost on training queries $\bx_\text{tr}\in\mathcal{X}_\text{tr}$. The second term is the variance of the per query zero mean noise $\epsilon^{(m)}(\bx)$ associated with the actual objective value $U^{(m)}(\bx)$. For the first term, 
\begin{align*}
    \mathbb{E}\big[(\overline{U}^{(m)}(\bx) - \widehat{U}^{(m)}(\bx))^2\big] &= \mathbb{E}[(\overline{U}^{(m)}(\bx) - \mathbb{E}(\widehat{U}^{(m)}(\bx)) + \mathbb{E}(\widehat{U}^{(m)}(\bx)) - \widehat{U}^{(m)}(\bx))^2]\\
    &= \mathbb{E}\big[ \big(\overline{U}^{(m)}(\bx) - \mathbb{E}(\widehat{U}^{(m)}(\bx))\big)^2 \big] + \mathbb{E}\big[ \big(\mathbb{E}(\widehat{U}^{(m)}(\bx)) - \widehat{U}^{(m)}(\bx)\big)^2 \big] \\
    & \qquad + 2 \,\mathbb{E} \big[ \big(\overline{U}^{(m)}(\bx) - \mathbb{E}(\widehat{U}^{(m)}(\bx))\big) \cdot \big(\mathbb{E}(\widehat{U}^{(m)}(\bx)) - \widehat{U}^{(m)}(\bx)\big) \big]
\end{align*}
After expanding and rearranging, the above expression reduces to 
\begin{align}\label{eq:bias_plus_variance}
    \mathbb{E}\big[\overline{U}^{(m)}(\bx) - \widehat{U}^{(m)}(\bx))^2\big] &= \big[\overline{U}^{(m)}(\bx) - \mathbb{E}(\widehat{U}^{(m)}(\bx)) \big]^2 + \mathbb{E}\big[ \big(\widehat{U}^{(m)}(\bx) - \mathbb{E}(\widehat{U}^{(m)}(\bx))\big)^2 \big]\\
     &= \text{Bias}^2(\widehat{U}^{(m)}(\bx)) + \text{Var}(\widehat{U}^{(m)}(\bx))\nonumber
\end{align}

Using the general representation for nonparametric routers $\widehat{U}^{(m)}(\bx) = \sum_i w_i(\bx) V^{(m)}_i$, 
\begin{align*}
    \text{Bias}^2(\widehat{U}^{(m)}(\bx)) &= \big[\overline{U}^{(m)}(\bx) - \mathbb{E}(\widehat{U}^{(m)}(\bx)) \big]^2  = \big[\overline{U}^{(m)}(\bx) - \mathbb{E}\Big(\sum_i w_i(\bx) \; V^{(m)}_i\Big) \big]^2 \\
    &= \big[U^{(m)}(\bx) - \sum_i w_i(\bx) \; \mathbb{E} [V^{(m)}_i] \big]^2 
\end{align*}
Using \cref{assumption:unbiased_centroid}, alongwith Jensen's inequality and local smoothness through \cref{assumption:smooth_objective}
\begin{align*}
    \text{Bias}^2(\widehat{U}^{(m)}(\bx))&= \Big[ \sum_i w_i(\bx) \big( \overline{U}^{(m)}(\bx) - \overline{U}^{(m)}(\br_i)\big)\Big]^2\\
    &\leq \sum_{i} w_i(\bx) \big( \overline{U}^{(m)}(\bx) - \overline{U}^{(m)}(\br_i) \big)^2 \\
    &\leq \sum_i w_i(\bx) \; L^2 \; d^2(\bx,\br_i)
\end{align*}
Similarly for the variance term $\text{Var}[\widehat{U}^{(m)}(\bx)]$,
\begin{align*}
    \text{Var}(\widehat{U}^{(m)}(\bx)) &= \mathbb{E}\big[ \big(\widehat{U}^{(m)}(\bx) - \mathbb{E}(\widehat{U}^{(m)}(\bx))\big)^2 \big] = \mathbb{E}\Big[ \sum_{i} w_i(\bx) \big( V^{(m)}_i - \mathbb{E}[V^{(m)}_i] \big) \Big]^2 \\
    &= \sum_{i} w_i^2(\bx) \; \text{Var}[V^{(m)}_i]
\end{align*}

For clustering based algorithms, $\text{Var}[V^{(m)}_i] = \mathbb{E}\Big[ \frac{1}{n_i^2}\sum_{\bx_\text{tr}\in c_i} (\epsilon^{(m)}(\bx_\text{tr}))^2 \Big] = \frac{1}{n_i^2}\sum_{\bx_\text{tr}\in c_i} \text{Var}[\epsilon^{(m)}(\bx_\text{tr})] \nonumber$ for cluster $c_i$ with $n_i$ number of training queries. For nearest neighbor based routers, $\text{Var}[V^{(m)}_i] = \text{Var}[\epsilon^{(m)}(\bx_\text{tr})]$ where $\bx_\text{tr}$ is the training query associated to reference element $i\in |\mathcal{I}|$.

Rewriting the squared error \cref{eq:squared_error}, and the bias and variance \cref{eq:bias_plus_variance} for the model objective value estimator at any test query $\bx$, we have
\begin{align*}
    \text{Error} (\widehat{U}^{(m)}(\bx), U^{(m)}(\bx)) &= \text{Bias}^2(\widehat{U}^{(m)}(\bx)) + \text{Var}(\widehat{U}^{(m)}(\bx)) + \text{Var}[\epsilon^{(m)}(\bx)]\\
    &\leq  \sum_i w_i(\bx) \; L^2 \; d^2(\bx,\br_i) + \sum_{i} w_i^2(\bx) \; \text{Var}[V^{(m)}_i] + \text{Var}[\epsilon^{(m)}(\bx)]
\end{align*}
\paragraph{Remarks:}  For choice of $\bw(\bx) = [w_1(\bx), \cdots, w_{|\mathcal{I}|}(\bx)]$ as the initial least variance prior $\bp(\bx) = [p_1(\bx), \cdots, p_{|\mathcal{I}|}(\bx)]$ in \cref{subsec:min_var_prior}, the variance term $\sum_{i} p_i^2(\bx) \; \text{Var}[V^{(m)}_i]$ is minimized and any other $\bw(\bx)\neq \bp(\bx)$ strictly increases variance. To reduce to bias term $\sum_i w_i(\bx) \; L^2 \; d^2(\bx,\br_i)$, we selectively prioritize closer reference elements to the test query $\bx$ by proximity based tilting. Choosing $w_i(\bx) \propto p_i(\bx) \exp(-\phi_i(\bx)/\tau)$ where proximity penalty $\phi_i(\bx)$ is an increasing function of the distance $d(\bx, \br_i)$ between test query $\bx$ and reference element $i$. Any super-logarithmic choice of proximity penalty leads to a decaying bias term in $d(\bx, \br_i)$, eventually leading to a balanced tradeoff for the squared error term \cref{eq:squared_error}. Note that the last term $\text{Var}[\epsilon^{(m)}(\bx)]$ captures the unavoidable noise in generated response for query $\bx$ by model $m$.

\section{Effective Sample Size} \label{app:effective_sample_size}
For any unbiased aggregation scheme over observed samples, the effective sample size corresponds to size of an unweighted sample average that would yield the variance reduction obtained through a weighted aggregation. For a general set of samples $\{ x_i=(\mu + \varepsilon_i): i \in [n] \}$, with zero mean uniform variance noise $\mathbb{E}[\varepsilon_i^2] = \sigma^2_i$, any unbiased estimate of the form 
\begin{equation*}
    \widehat{\mu} = \sum_{i\in [n]} w_i \;x_i \;, \qquad w_i\geq 0, \;\;\sum_i w_i = 1.
\end{equation*}
has variance $\sum_i w_i^2 \;\sigma_i^2$. The effective sample size for such an estimator is:
\begin{equation*}
    \text{ESS}(\widehat{\mu}) = \frac{\sum w_i^2}{(\sum w_i)^2} = \sum w_i^2
\end{equation*}
So $\text{Var}(\widehat{\mu}) = \sigma^2/ \text{ESS}(\widehat{\mu})$. Any deviation from uniform averaging weights over the same $[n]$ points yields a strictly higher variance estimate.\\

As found in \Cref{app:min_var_prior}, the optimal estimator in heteroscedastic noise is $w_i \propto 1/\sigma_i^2$, and the variance of the corresponding estimator is $\text{Var}(\widehat{\mu}) = 1/ \sum_i \sigma_i^{-2}$. Any deviation from these aggregation weights in the heteroscedastic setting yields a strictly higher variance estimate. For our analysis of techniques in $k$NN based aggregation (\Cref{tab:knn_proxrouter}), we utilize the closest $k=100$ nearest neighbors of any test query to ensure fair comparison which inevitably leads to a higher variance in $k$NN-Prox causing a slight routing performance drop on inlier queries. Practically, $k$NN-Prox leads to a smaller effective sample size over the closest $k$ queries, which can be overcome by choosing a larger set of neighboring queries.

\section{ProxRouter Extended Results}\label{app:extended_results}
\subsection{Detailed Results for \Cref{sec:experiments}}

In this section, we present details on results already presented in the main segment of \Cref{sec:experiments}.

\begin{figure}[htbp]
  \centering

  \begin{subfigure}{\textwidth}
    \centering
    \begin{minipage}{0.32\textwidth}
      \includegraphics[width=\linewidth]{aistats/images/router_performance/gsmk_svamp/knn/ood/pareto_accuracy_cost.pdf}
    \end{minipage}\hfill
    \begin{minipage}{0.32\textwidth}
      \includegraphics[width=\linewidth]{aistats/images/router_performance/gsmk_svamp/knn/ood/accuracy_vs_lambda.pdf}
    \end{minipage}\hfill
    \begin{minipage}{0.32\textwidth}
      \includegraphics[width=\linewidth]{aistats/images/router_performance/gsmk_svamp/knn/ood/cost_vs_lambda.pdf}
    \end{minipage}
    \caption{Routing Performance on outlier test queries.}
  \end{subfigure}

  \vspace{0.75em}

  \begin{subfigure}{\textwidth}
    \centering
    \begin{minipage}{0.32\textwidth}
      \includegraphics[width=\linewidth]{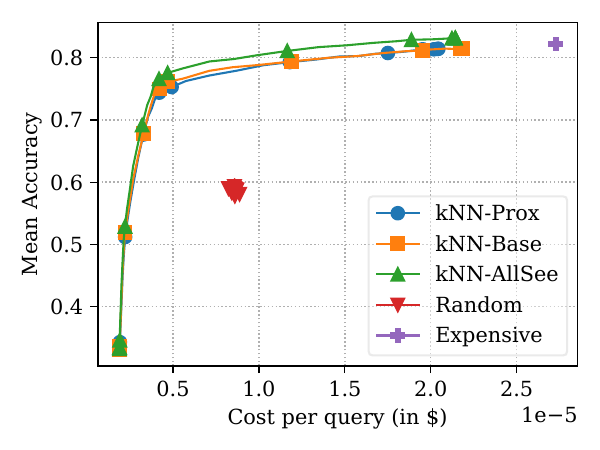}
    \end{minipage}\hfill
    \begin{minipage}{0.32\textwidth}
      \includegraphics[width=\linewidth]{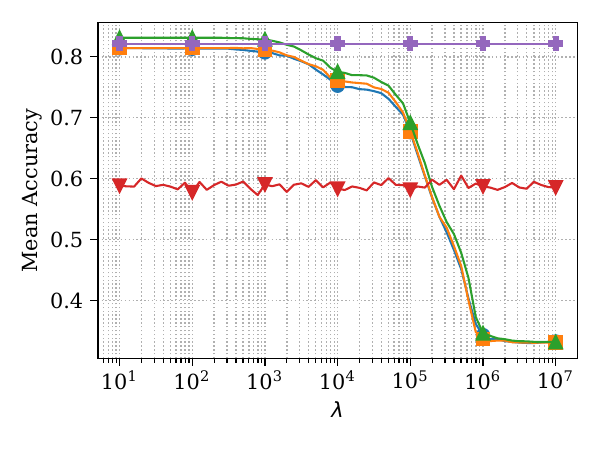}
    \end{minipage}\hfill
    \begin{minipage}{0.32\textwidth}
      \includegraphics[width=\linewidth]{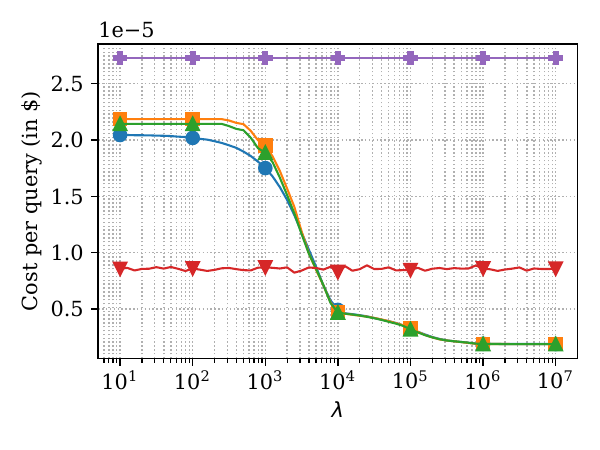}
    \end{minipage}
    \caption{Routing Performance on inlier test queries.}
  \end{subfigure}

  \vspace{0.75em}

  \begin{subfigure}{\textwidth}
    \centering
    \begin{minipage}{0.32\textwidth}
      \includegraphics[width=\linewidth]{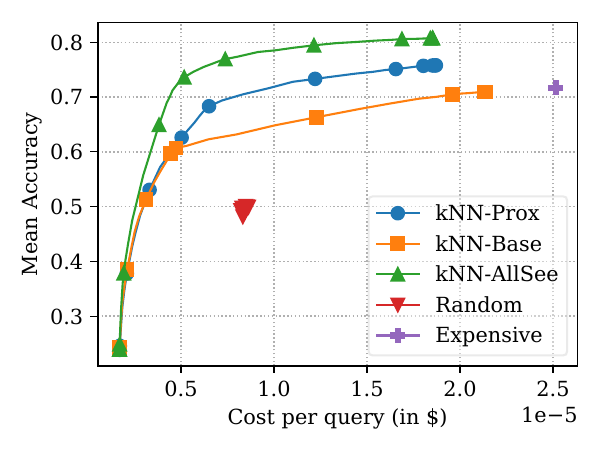}
    \end{minipage}\hfill
    \begin{minipage}{0.32\textwidth}
      \includegraphics[width=\linewidth]{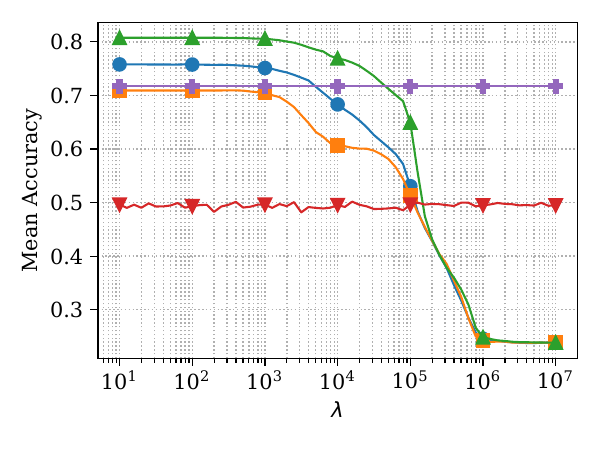}
    \end{minipage}\hfill
    \begin{minipage}{0.32\textwidth}
      \includegraphics[width=\linewidth]{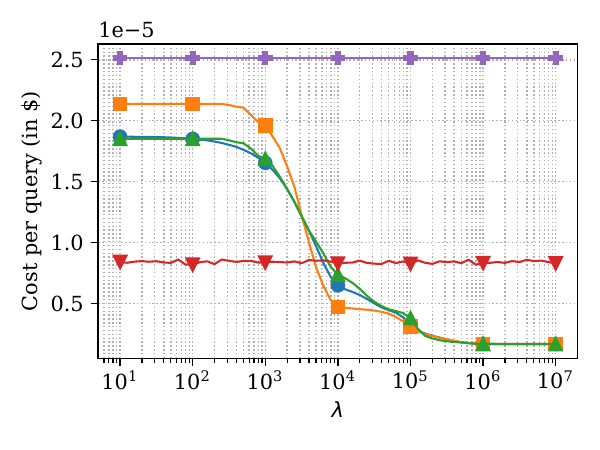}
    \end{minipage}
    \caption{Routing Performance on all test queries.}
  \end{subfigure}

  \caption{$k$NN-Prox performance on GSM8K+SVAMP outlier tasks,
  \Cref{tab:knn_proxrouter}. Note that outlier performance is greatly improved while preserving inlier query routing performance.}
  \label{fig:gsm8k_svamp_full_performance}
\end{figure}

\Cref{fig:gsm8k_svamp_full_performance} denotes the performance of $k$NN-Prox as presented in \Cref{tab:knn_proxrouter}. 

\begin{figure}[htbp]
  \centering

  \begin{subfigure}{\textwidth}
    \centering
    \begin{minipage}{0.32\textwidth}
      \includegraphics[width=\linewidth]{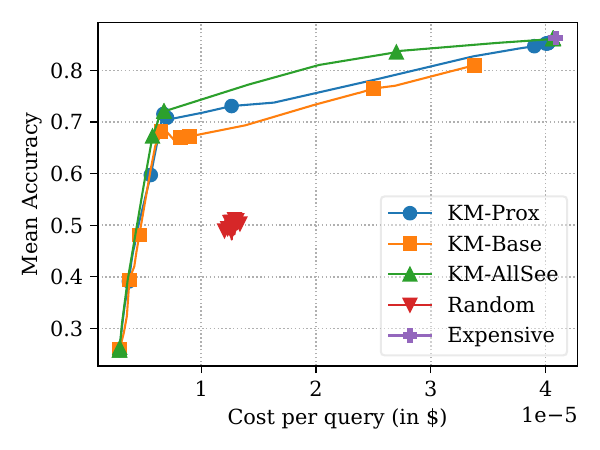}
    \end{minipage}\hfill
    \begin{minipage}{0.32\textwidth}
      \includegraphics[width=\linewidth]{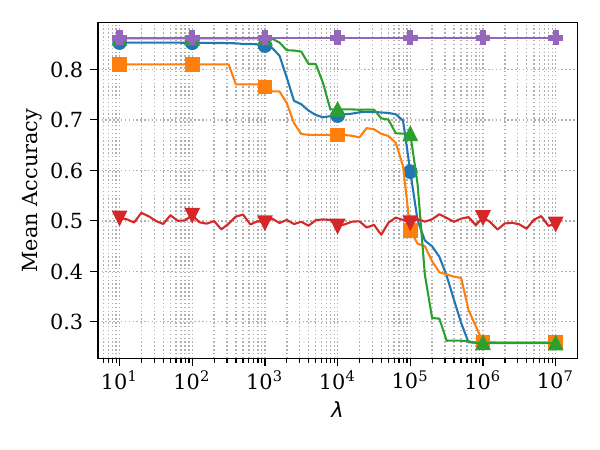}
    \end{minipage}\hfill
    \begin{minipage}{0.32\textwidth}
      \includegraphics[width=\linewidth]{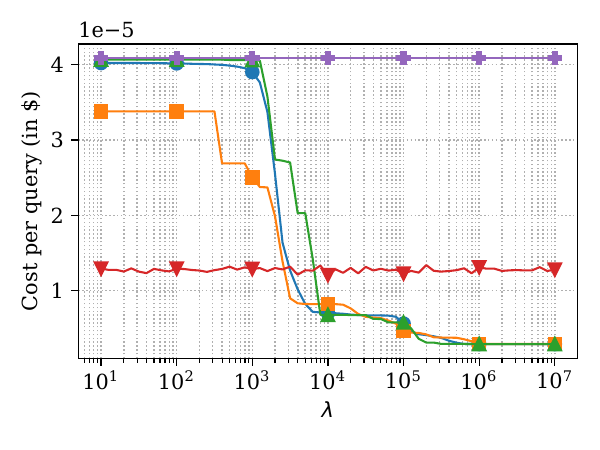}
    \end{minipage}
    \caption{Routing Performance on outlier test queries.}
  \end{subfigure}

  \vspace{0.75em}

  \begin{subfigure}{\textwidth}
    \centering
    \begin{minipage}{0.32\textwidth}
      \includegraphics[width=\linewidth]{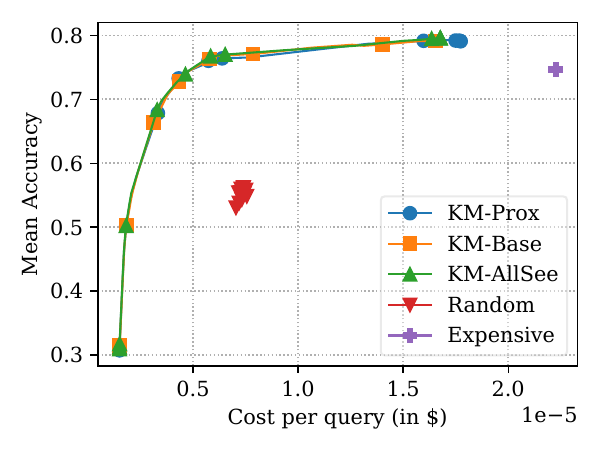}
    \end{minipage}\hfill
    \begin{minipage}{0.32\textwidth}
      \includegraphics[width=\linewidth]{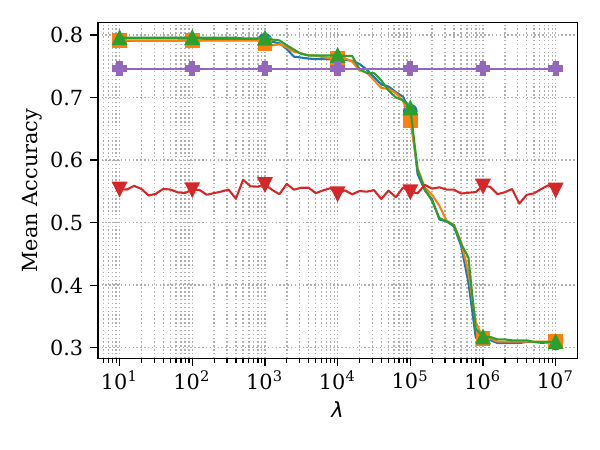}
    \end{minipage}\hfill
    \begin{minipage}{0.32\textwidth}
      \includegraphics[width=\linewidth]{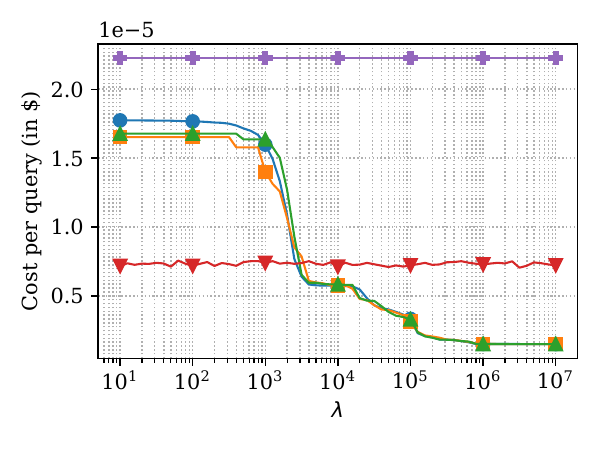}
    \end{minipage}
    \caption{Routing Performance on inlier test queries.}
  \end{subfigure}

  \vspace{0.75em}

  \begin{subfigure}{\textwidth}
    \centering
    \begin{minipage}{0.32\textwidth}
      \includegraphics[width=\linewidth]{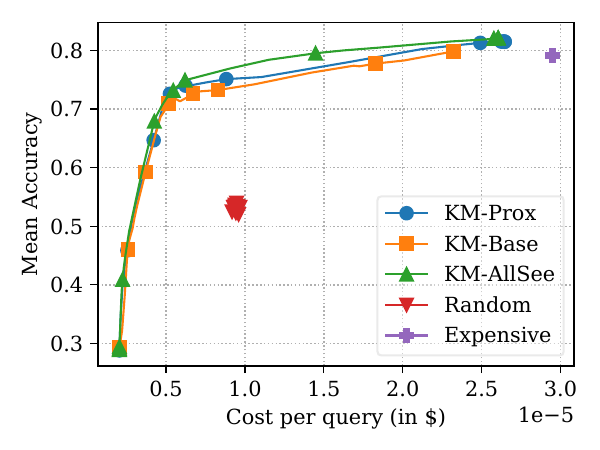}
    \end{minipage}\hfill
    \begin{minipage}{0.32\textwidth}
      \includegraphics[width=\linewidth]{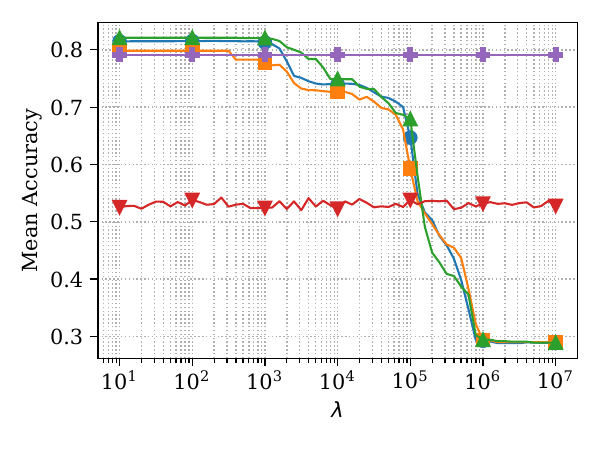}
    \end{minipage}\hfill
    \begin{minipage}{0.32\textwidth}
      \includegraphics[width=\linewidth]{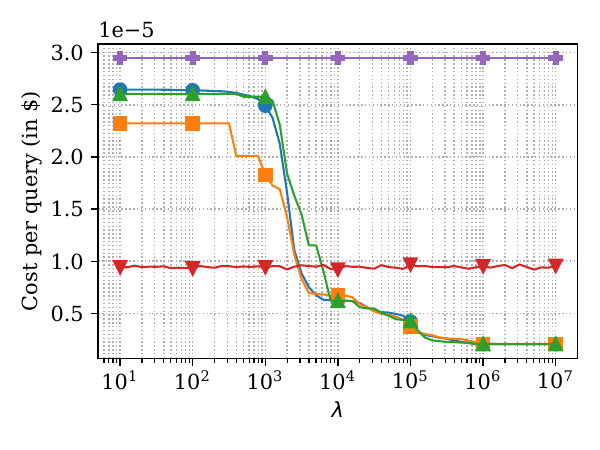}
    \end{minipage}
    \caption{Routing Performance on all test queries.}
  \end{subfigure}

  \caption{$K$M-Prox performance on MedQA+Hellaswag outlier tasks,
  \Cref{tab:kmeans_proxrouter}. Note that outlier performance is greatly improved while preserving inlier query routing performance.}
  \label{fig:medqa_hellaswag_full_performance}
\end{figure}

\begin{figure}[htbp]
  \centering

  \begin{subfigure}{\textwidth}
    \centering
    \begin{minipage}{0.32\textwidth}
      \includegraphics[width=\linewidth]{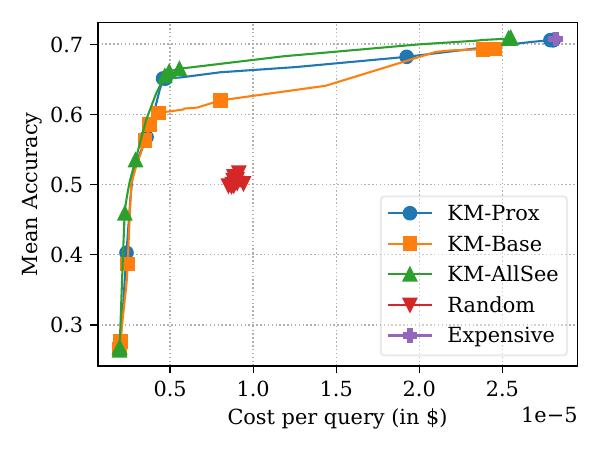}
    \end{minipage}\hfill
    \begin{minipage}{0.32\textwidth}
      \includegraphics[width=\linewidth]{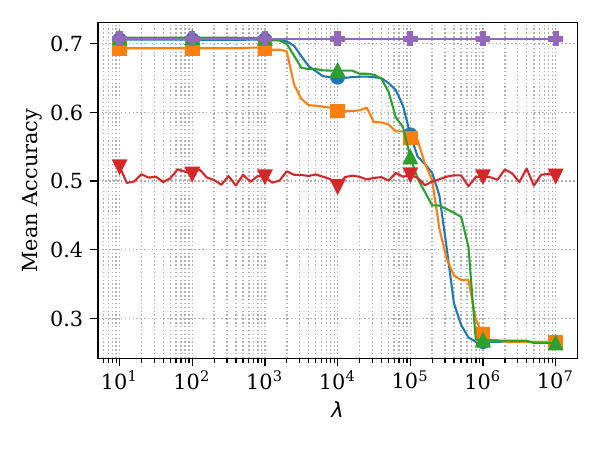}
    \end{minipage}\hfill
    \begin{minipage}{0.32\textwidth}
      \includegraphics[width=\linewidth]{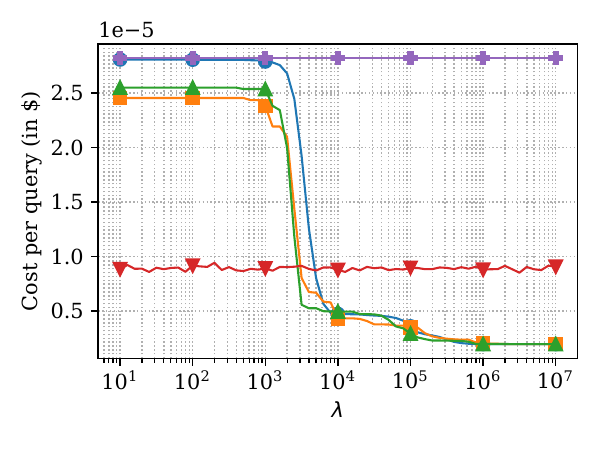}
    \end{minipage}
    \caption{Routing Performance on outlier test queries.}
  \end{subfigure}

  \vspace{0.75em}

  \begin{subfigure}{\textwidth}
    \centering
    \begin{minipage}{0.32\textwidth}
      \includegraphics[width=\linewidth]{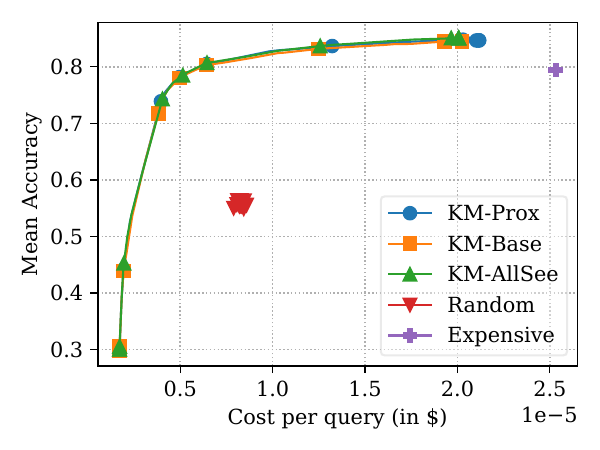}
    \end{minipage}\hfill
    \begin{minipage}{0.32\textwidth}
      \includegraphics[width=\linewidth]{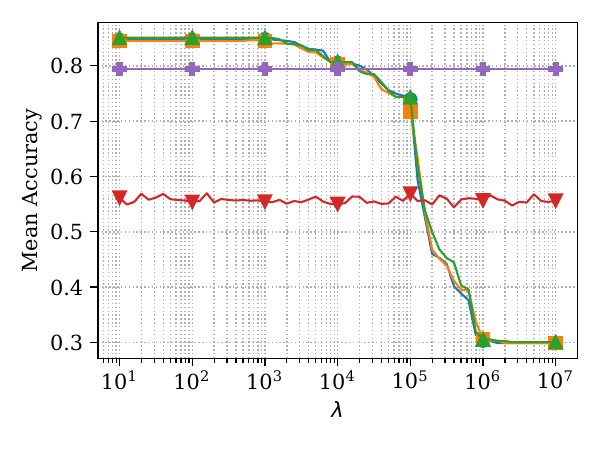}
    \end{minipage}\hfill
    \begin{minipage}{0.32\textwidth}
      \includegraphics[width=\linewidth]{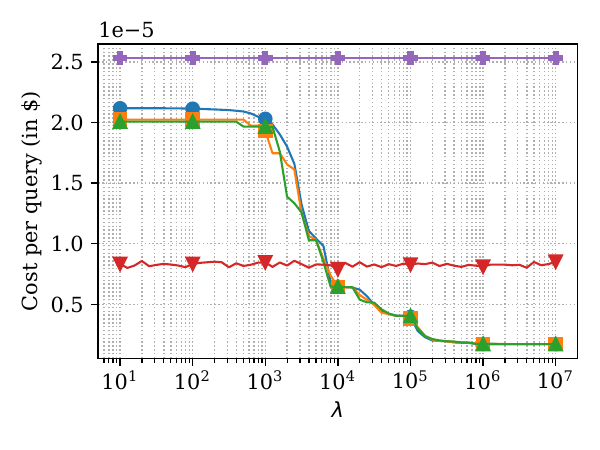}
    \end{minipage}
    \caption{Routing Performance on inlier test queries.}
  \end{subfigure}

  \vspace{0.75em}

  \begin{subfigure}{\textwidth}
    \centering
    \begin{minipage}{0.32\textwidth}
      \includegraphics[width=\linewidth]{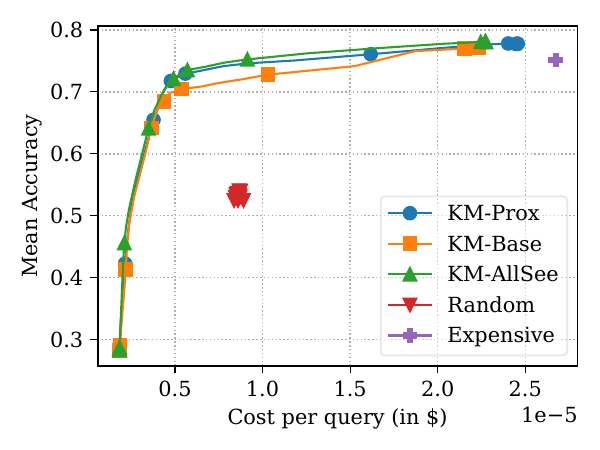}
    \end{minipage}\hfill
    \begin{minipage}{0.32\textwidth}
      \includegraphics[width=\linewidth]{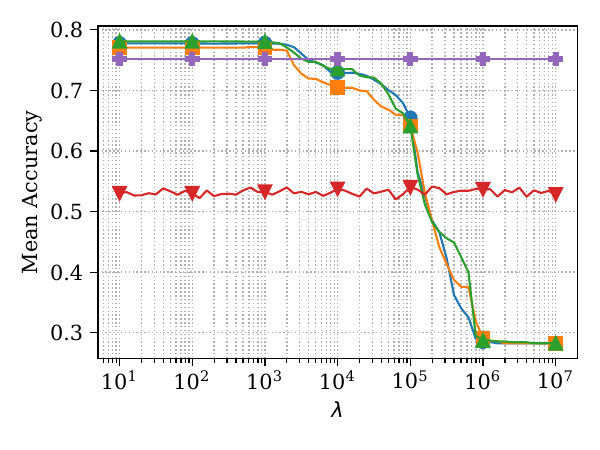}
    \end{minipage}\hfill
    \begin{minipage}{0.32\textwidth}
      \includegraphics[width=\linewidth]{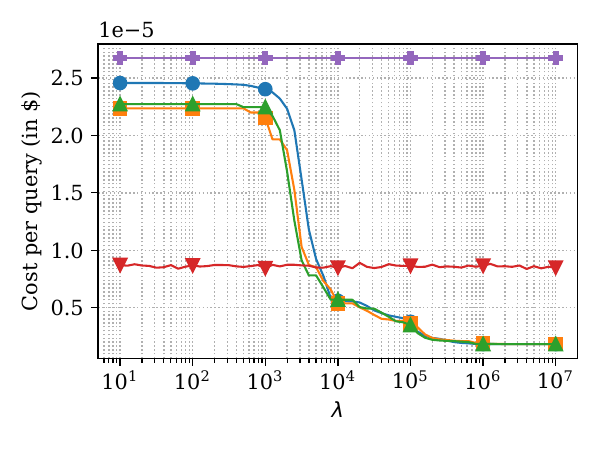}
    \end{minipage}
    \caption{Routing Performance on all test queries.}
  \end{subfigure}

  \caption{$K$M-Prox performance on LogiQA+CommonSenseQA+BBH(BoolEx) outlier tasks,
  \Cref{tab:kmeans_proxrouter}. Note that outlier performance is greatly improved while preserving inlier query routing performance.}
  \label{fig:logiqa_csqa_bbh_full_performance}
\end{figure}

\Cref{fig:medqa_hellaswag_full_performance,fig:logiqa_csqa_bbh_full_performance} denote the performance of $K$M-Prox as presented in \Cref{tab:kmeans_proxrouter}.

\subsection{Ablations for Num clusters ($K$Means Router), Num Neighbors ($k$NN Router)}
In this section, we present the routing performance values for different number of clusters $K$ for clustering based routers (\Cref{tab:ablation_km_proxrouter}), and for different values of nearest neighbors $k$ for nearest neighbor based routers (\Cref{tab:ablation_knn_proxrouter}).

\begin{table}[htb]
\centering
\small
\caption{\small Performance (AUC normalized) of $K$M-Prox (ProxRouter) vs. $K$M-Base for multiple values of number of clusters $K$. Upper bound denoted by the full knowledge router $K$M-AllSee.}
\label{tab:ablation_km_proxrouter}
\setlength{\tabcolsep}{2pt}
\begin{tabular}{@{}cccccc@{}}
\toprule
\multicolumn{1}{c}{\multirow{2}{*}{$K$}} &
\multicolumn{1}{c}{\multirow{2}{*}{\shortstack{Outlier Tasks}}} &
\multicolumn{1}{c}{\multirow{2}{*}{Split}} &
\multicolumn{2}{c}{Routing method} &
\multicolumn{1}{c}{Upper Bound} \\
\cmidrule(lr){4-5}\cmidrule(lr){6-6}
 &  &  & $K$M-Base & $K$M-Prox & $K$M-AllSee \\
\midrule

\multicolumn{1}{c}{\multirow{6}{*}{16}} &
  \multirow{3}{*}{\shortstack{Hellaswag,\\ MedQA}} &
  Outlier & 71.24\% & 75.11\% & 78.04\% \\
 &  & Inlier  & 74.77\% & 74.82\% & 75.16\% \\
 &  & Overall & 73.64\% & 75.24\% & 76.59\% \\
  \addlinespace
 & \multirow{3}{*}{\shortstack{LogiQA,\\ CSQA, BBH}} &
  Outlier & 65.60\% & 66.22\% & 67.07\% \\
 &  & Inlier  & 79.97\% & 80.05\% & 79.90\% \\
 &  & Overall & 73.07\% & 73.54\% & 73.96\% \\
\midrule 

\multicolumn{1}{c}{\multirow{6}{*}{32}} &
  \multirow{3}{*}{\shortstack{Hellaswag,\\ MedQA}} &
  Outlier & 70.68\% & 74.88\% & 78.36\% \\
 &  & Inlier  & 74.62\% & 74.86\% & 74.63\% \\
 &  & Overall & 73.04\% & 75.12\% & 74.87\% \\
 \addlinespace
 & \multirow{3}{*}{\shortstack{LogiQA,\\ CSQA, BBH}} &
  Outlier & 63.39\% & 66.18\% & 67.25\% \\
 &  & Inlier  & 79.35\% & 79.92\% & 79.94\% \\
 &  & Overall & 71.61\% & 73.46\% & 73.88\% \\
\midrule

\multicolumn{1}{c}{\multirow{6}{*}{64}} &
  \multirow{3}{*}{\shortstack{Hellaswag,\\ MedQA}} &
  Outlier & 70.65\% & 74.72\% & 78.56\% \\
 &  & Inlier  & 74.27\% & 74.70\% & 75.38\% \\
 &  & Overall & 72.96\% & 74.96\% & 76.87\% \\
  \addlinespace
 & \multirow{3}{*}{\shortstack{LogiQA,\\ CSQA, BBH}} &
  Outlier & 64.02\% & 66.12\% & 67.74\% \\
 &  & Inlier  & 78.71\% & 79.66\% & 80.15\% \\
 &  & Overall & 71.70\% & 73.32\% & 74.24\% \\

\bottomrule
\end{tabular}
\end{table}

\begin{table}[htb]
\centering
\small
\caption{\small Performance (AUC normalized) of $k$NN-Prox (ProxRouter) vs. $k$NN-Base for multiple values of nearest neighbors $k$. Upper bound denoted by the full knowledge router $k$NN-AllSee.}
\label{tab:ablation_knn_proxrouter}
\setlength{\tabcolsep}{2pt}
\begin{tabular}{@{}cccccc@{}}
\toprule
\multicolumn{1}{c}{\multirow{2}{*}{$k$}} &
\multicolumn{1}{c}{\multirow{2}{*}{\shortstack{Outlier Tasks}}} &
\multicolumn{1}{c}{\multirow{2}{*}{Split}} &
\multicolumn{2}{c}{Routing method} &
\multicolumn{1}{c}{Upper Bound} \\
\cmidrule(lr){4-5}\cmidrule(lr){6-6}
 &  &  & $k$NN-Base & $k$NN-Prox & $k$NN-AllSee \\
\midrule 

\multicolumn{1}{c}{\multirow{3}{*}{50}} &
  \multirow{3}{*}{\shortstack{GSM8k,\\ SVAMP}} &
  Outlier & 44.52\% & 47.27\% & 61.60\% \\
 &  & Inlier  & 76.46\% & 76.23\% & 79.66\% \\
 &  & Overall & 66.00\% & 68.28\% & 75.00\% \\
\midrule

\multicolumn{1}{c}{\multirow{3}{*}{100}} &
  \multirow{3}{*}{\shortstack{GSM8k,\\ SVAMP}} &
  Outlier & 38.55\% & 46.64\% & 60.77\% \\
 &  & Inlier  & 77.51\% & 76.96\% & 79.11\% \\
 &  & Overall & 63.98\% & 68.12\% & 74.60\% \\
\midrule 

\multicolumn{1}{c}{\multirow{3}{*}{200}} &
  \multirow{3}{*}{\shortstack{GSM8k,\\ SVAMP}} &
  Outlier & 38.83\% & 46.10\% & 59.60\% \\
 &  & Inlier  & 77.83\% & 77.08\% & 78.85\% \\
 &  & Overall & 64.15\% & 67.61\% & 74.35\% \\
\bottomrule
\end{tabular}
\end{table}

\subsection{Parametric and Nonparametric Router Comparison}
\begin{figure}
    \centering
    \includegraphics[width=0.4\linewidth]{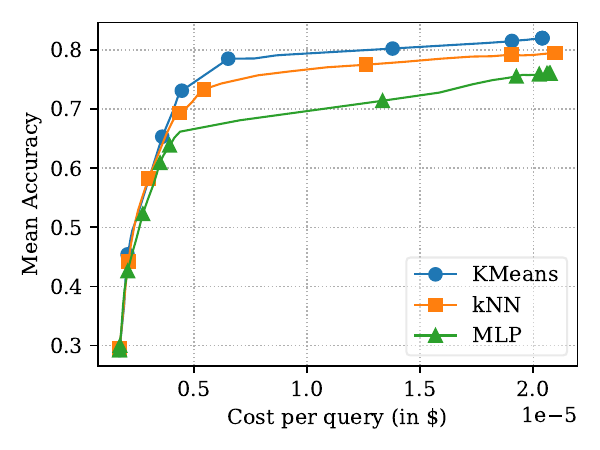}
    \caption{Nonparametric routers match, or even outperform learned parametric approaches such as MLP based routers. We utilize a two head MLP layer for predicting accuracies and costs of model responses respectively, and compare routing performance with existing $k$NN and $K$Means approaches. Same experimental setting used from \Cref{tab:model_performance_grouped}.}
    \label{fig:mlp_vs_nonparametric}
\end{figure}

We compare nonparametric approaches with a parametric approach utilizing learned neural network router (MLP layers) on top of the query sentence encodings. Our MLP router utilizes a two head output where each head's output dimension is the number of models in the pool ($|\mathcal{M}|$), one head predicting accuracy and the other head predicting cost of response for all models. Model selection is done through the same approach, where \cref{eq:routing_objective} is now calculated using these accuracy and cost estimates for each model for a given value of $\lambda$. We also reiterate that MLP router needs full retraining to accommodate a change in the model pool or addition of newer tasks in the training set, unlike $K$Means or $k$NN based routers which can naturally accommodate such changes in an incremental manner. We use the same experimental setting in \Cref{tab:model_performance_grouped}, with first half tasks as outliers and other half as inliers. \Cref{fig:mlp_vs_nonparametric} presents our comparison of parametric (MLP Router) and nonparametric ($k$NN and $K$Means routers) query routing approaches, highlighting that nonparametric methods match or even outperform complicated parametric routers. 

\section{Reproducibility Checklist}
We generate responses from LLMs for all prompts using a fixed seed (\texttt{42}). The query encodings generated from sentence encoders are deterministic, and so are the encodings of training queries. We run $K$Means clustering algorithm using a constant seed (\texttt{42}) for initialization of cluster centers, but we observed almost indistinguishable results for other seeds as well, indicating stable clustering. Nearest neighbor based schemes are fully deterministic due to the deterministic nature of encodings, hence we present the results directly. All train-test splits are also created deterministically with the same choice of random seed. For additional analysis such as delay modeling, we repeat our experiments over five different seeds $(7,42,99,1234,2024)$ and present the average results. We furthermore pin the experimental setting in delay analysis by choosing one CPU thread for avoiding multi-threading jitters and dynamic BLAS thread allocation artifacts for reflecting true time complexity of our approach. 

\end{document}